\documentclass[10pt,journal,compsoc]{IEEEtran}

\usepackage{booktabs} 
\usepackage{tabularx}
\usepackage{graphicx}
\usepackage{url}
\usepackage{mdwlist}
\usepackage{subcaption}
\usepackage{hyperref}
\hypersetup{colorlinks=false,linkcolor=blue,urlcolor=blue,citecolor=red}
\usepackage{amsfonts}
\usepackage{amsmath}
\usepackage[labelfont=bf,textfont=bf]{caption}
\usepackage{multirow}
\usepackage{array}
\usepackage[linesnumbered,ruled]{algorithm2e}
\SetKwComment{Comment}{$\triangleright$\ }{}
\usepackage{xcolor}
\usepackage{amssymb}
\usepackage[flushleft]{threeparttable}
\usepackage{enumitem}
\usepackage{balance}

\newcommand{\ie}{\textit{i.e.}}
\newcommand{\eg}{\textit{e.g.}}
\newcommand{\etal}{\textit{et al.}}

\ifCLASSOPTIONcompsoc
  \usepackage[nocompress]{cite}
\else
  \usepackage{cite}
\fi

\begin{document}

\title{Context-Aware Visual Policy Network \\ for Fine-Grained Image Captioning}

\author{Zheng-Jun~Zha,~\IEEEmembership{Member,~IEEE}
        Daqing~Liu,
        Hanwang~Zhang,~\IEEEmembership{Member,~IEEE},
        Yongdong Zhang,~\IEEEmembership{Member,~IEEE},
        and~Feng~Wu,~\IEEEmembership{Fellow,~IEEE}
\IEEEcompsocitemizethanks{\IEEEcompsocthanksitem Z.-J. Zha, D. Liu, Y. Zhang and F. Wu are with the School of Information Science and Technology, University of Science and Technology of China.\protect\\
E-mail: zhazj@ustc.edu.cn
\IEEEcompsocthanksitem H. Zhang is with the School of Computer Science and Engineering, Nanyang Technological University.}
}

\markboth{IEEE Trans. Pattern Analysis and Machine Intelligence}%
{Zha \MakeLowercase{\textit{et al.}}: Context-Aware Visual Policy Network for Fine-Grained Image Captioning}

\IEEEtitleabstractindextext{
\begin{abstract}
With the maturity of visual detection techniques, we are more ambitious in describing visual content with open-vocabulary, fine-grained and free-form language, \textit{i.e.}, the task of image captioning. In particular, we are interested in generating longer, richer and more fine-grained sentences and paragraphs as image descriptions.
Image captioning can be translated to the task of sequential language prediction given visual content, where the output sequence forms natural language description with plausible grammar. However, existing image captioning methods focus only on language policy while not visual policy, and thus fail to capture visual context that is crucial for compositional reasoning such as object relationships (\textit{e.g.}, ``man riding horse'') and visual comparisons (\textit{e.g.}, ``small(er) cat''). This issue is especially severe when generating longer sequences such as a paragraph.
To fill the gap, we propose a Context-Aware Visual Policy network (CAVP) for fine-grained image-to-language generation: image sentence captioning and image paragraph captioning.
During captioning, CAVP explicitly considers the previous visual attentions as context and decides whether the context is used for the current word/sentence generation given the current visual attention.
Compared against the traditional visual attention mechanism that only fixes a single visual region at each step, CAVP can attend to complex visual compositions over time. The whole image captioning model --- CAVP and its subsequent language policy network --- can be efficiently optimized end-to-end by using an actor-critic policy gradient method. We have demonstrated the effectiveness of CAVP by state-of-the-art performances on MS-COCO and Stanford captioning datasets, using various metrics and sensible visualizations of qualitative visual context.
\end{abstract}

}
\maketitle
\IEEEdisplaynontitleabstractindextext
\IEEEpeerreviewmaketitle

\section{Introduction}
\begin{figure*}
	\begin{subfigure}{0.33\textwidth}
		\centering
		\includegraphics[width=\linewidth]{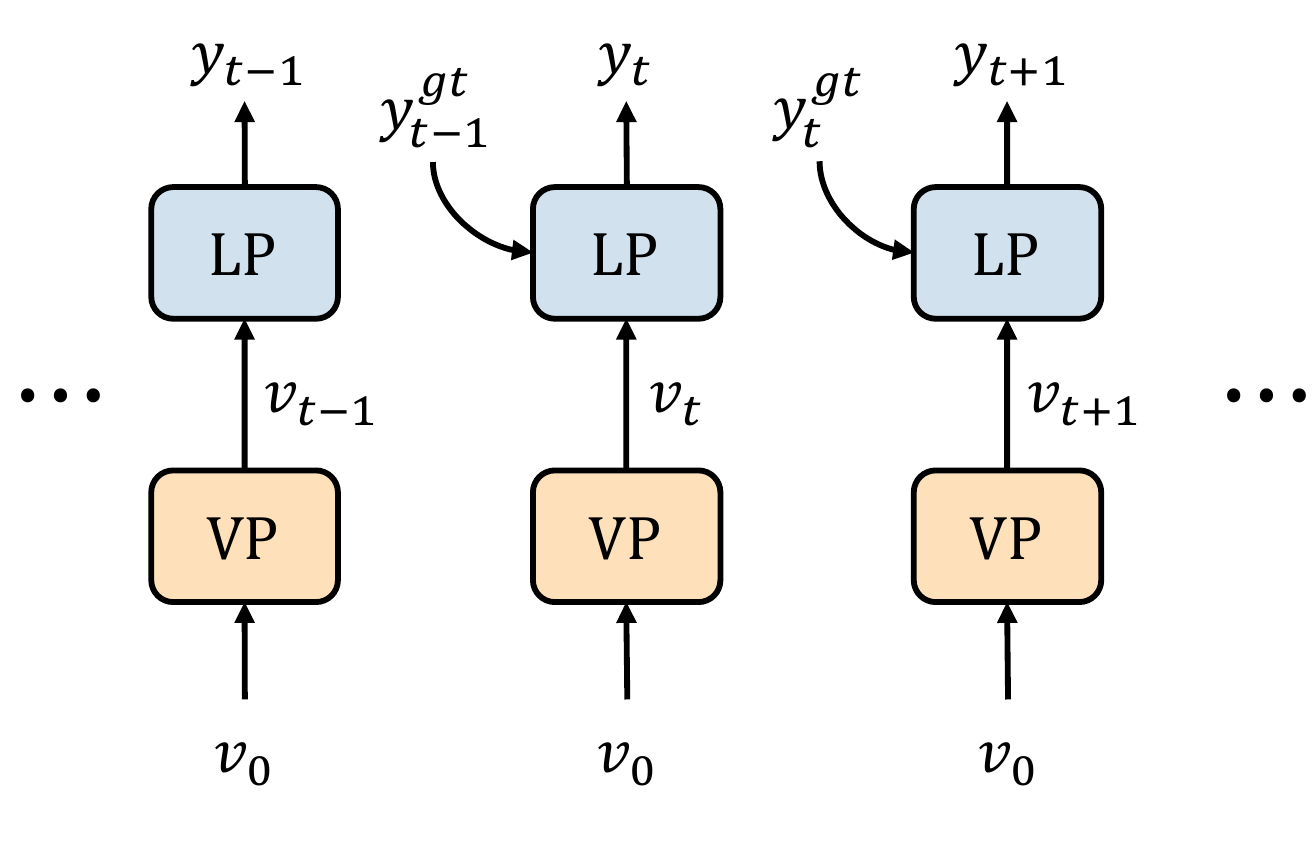}
		\caption{Traditional Framework}
        \label{fig:2a}
	\end{subfigure}
	\begin{subfigure}{0.33\textwidth}
		\centering
		\includegraphics[width=\linewidth]{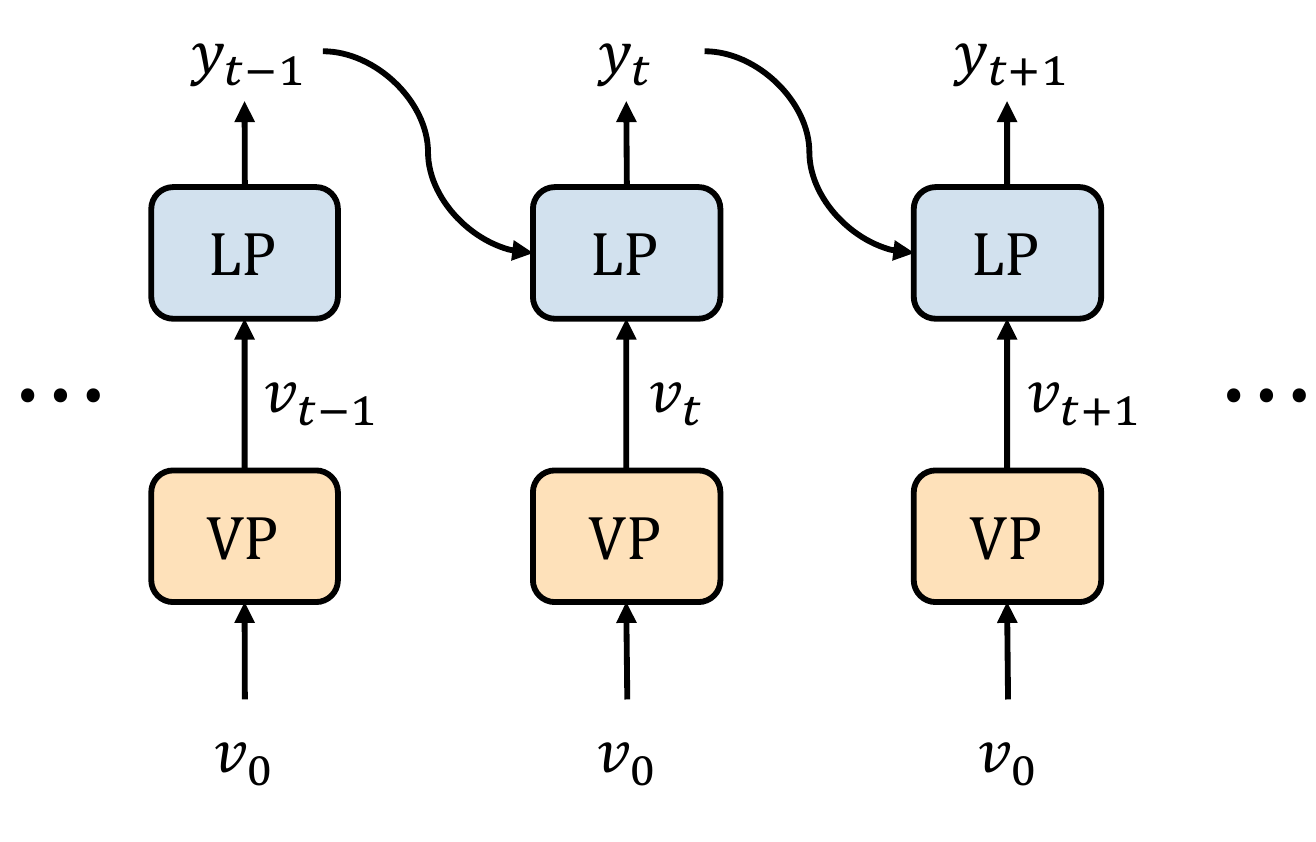}
		\caption{RL-based Framework}
        \label{fig:2b}
	\end{subfigure}
	\begin{subfigure}{0.33\textwidth}
		\centering
		\includegraphics[width=\linewidth]{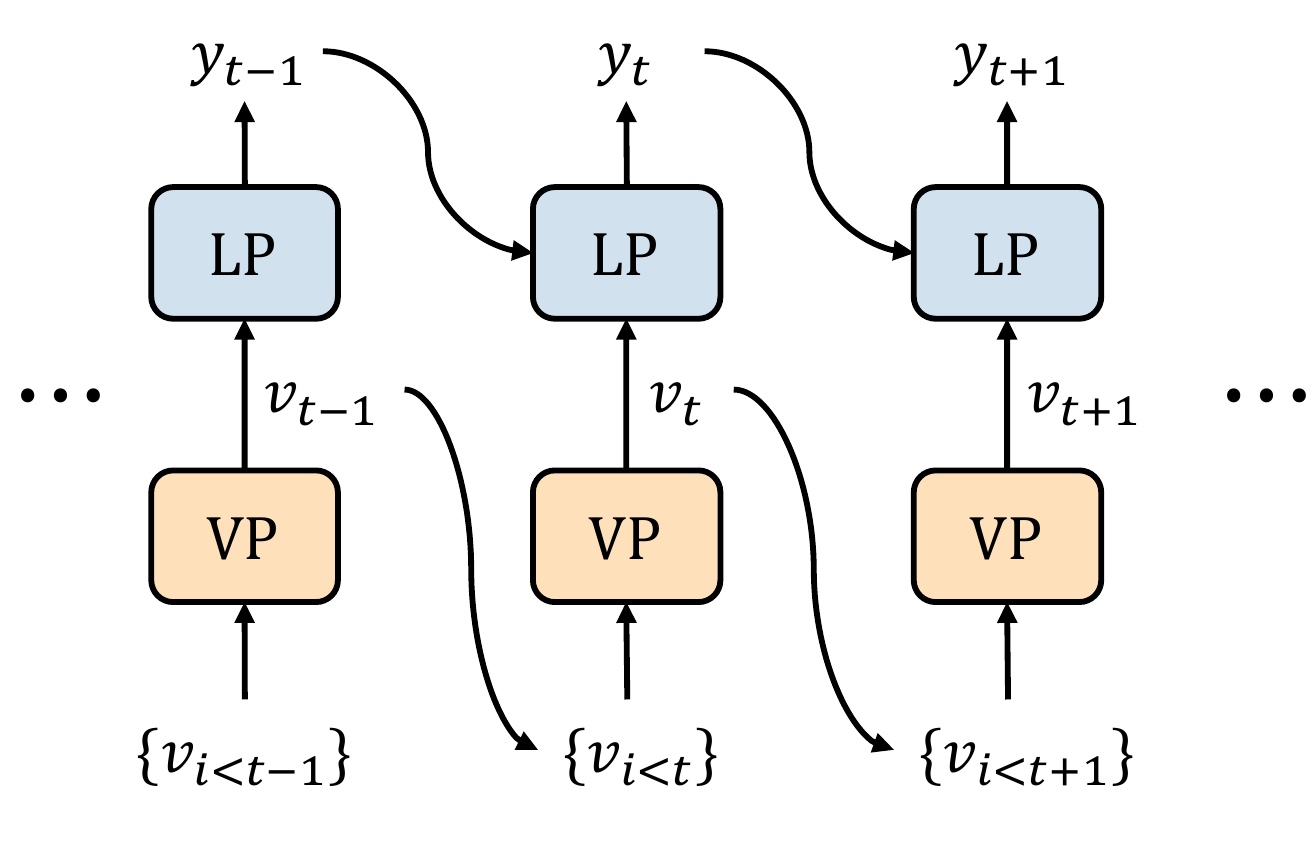}
		\caption{Our Framework}
        \label{fig:2c}
	\end{subfigure}
	\caption{The evolution of the encoder-decoder framework for image captioning. LP: language policy. VP: visual policy. $v_t$: visual feature at step $t$. $y_t$: predicted word at step $t$. $y^{gt}_t$: ground-truth word at step $t$. (a) The traditional framework focuses only on word prediction by exposing the ground-truth word $y^{gt}_{t-1}$ as input to step $t$ for language generation. (b) RL-based framework focuses on sequence training by directly feeding the predicted word $y_{t-1}$ to LP at step $t$. (c) Our proposed framework explicitly takes historical visual actions $\{v_{i < t}\}$ as visual context at step $t$.}
	\label{fig:2}
\end{figure*}

\begin{figure}
\centering
	\includegraphics[width=0.95\linewidth]{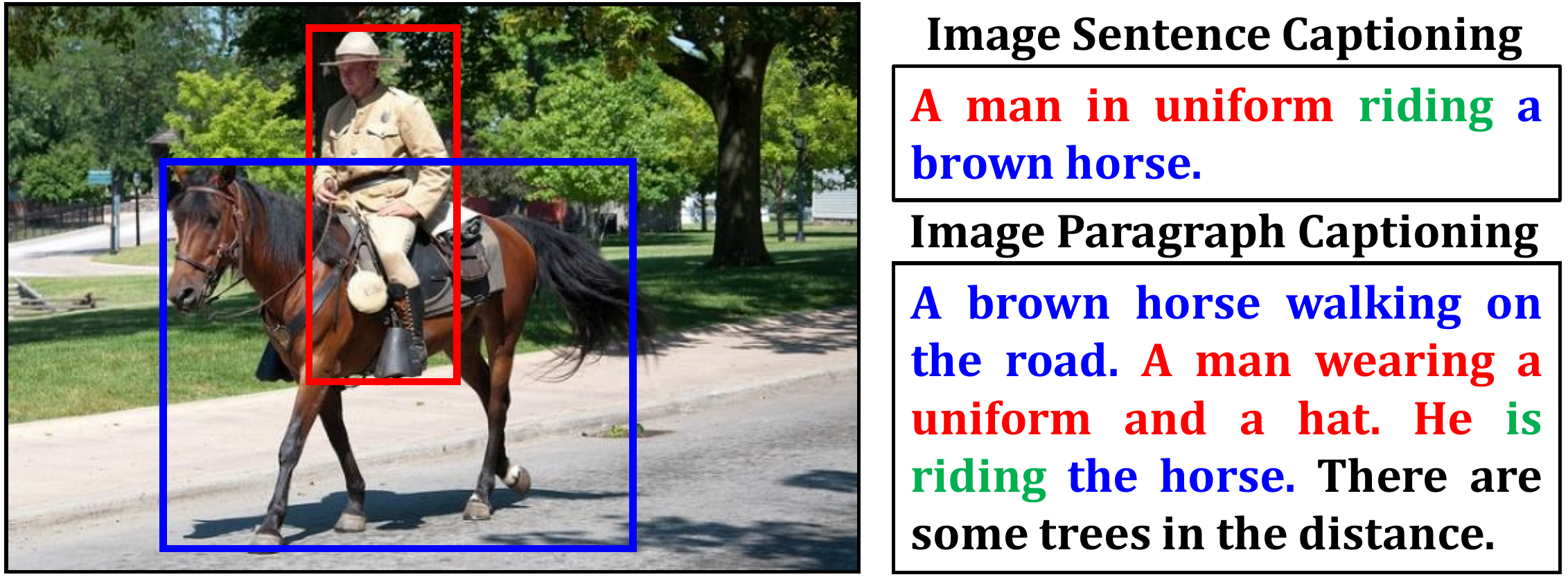}
    \caption{The intuition of using visual context in fine-grained image captioning. The proposed CAVP is the first RL-based image captioning model which incorporates visual context into sequential visual reasoning.}
    \label{fig:intuition}
    \vspace{-4mm}
\end{figure}

Vision and natural language machine comprehension --- the ever-lasting goal in Artificial Intelligence --- is rapidly evolving with the help of deep learning based AI technologies~\cite{hochreiter1997long, krizhevsky2012imagenet, ren2017deep, bahdanau2014neural}. The effective visual~\cite{krizhevsky2012imagenet, ren2017deep, ren2015faster} and textual representations~\cite{hochreiter1997long, bahdanau2014neural} empower computer vision systems to migrate from fixed-vocabulary, coarse-grained, and low-level visual analysis, \eg, image classification~\cite{ren2017deep} and object detection~\cite{ren2015faster}, to open-vocabulary, fine-grained, and high-level visual description, \eg, image captioning~\cite{vinyals2015show, xu2015show} and visual question answering~\cite{antol2015vqa}. The former has become relatively mature. However, the latter is still far from satisfactory, due to the lack of reasoning capability of deep neural networks~\cite{zhang2017interpretable}. Machine reasoning requires a series of complicated decisions, including inferring task-related context, identifying its efficacy for the current on-going task, as well as modeling the relationships between the context and task. How to build machines that can reason as humans is still a very challenging task~\cite{lake2017building}.

A prime example is image captioning --- the task describing images with natural language --- which demonstrates a machine's visual comprehension in terms of its ability of grounded natural language modeling~\cite{vinyals2015show, xu2015show}. In order for this AI-complete task~\cite{geman2015visual}, researchers have attempted to combine the most advanced computer vision (CV) techniques like object recognition~\cite{ren2015faster}, relationship detection~\cite{zhang2017visual}, and scene parsing~\cite{xu2017scene}, as well as the modern natural language processing (NLP) techniques such as language generative models~\cite{bahdanau2014neural,yu2017seqgan}.
In a nutshell, the CV-end acts as an encoder and the NLP-end plays as a decoder, translating from ``source'' image to ``target'' language. Such encoder-decoder architecture is trained using human-annotated image and sentence pairs in a fully-supervised way. The decoder is supervised to maximize the posterior probability of each ground-truth word given the previous ground-truth subsequence and ``source'' image. Unfortunately, due to the exponentially large search space of language compositions, recent studies have shown that such conventional supervised training tends to learn data bias but not machine reasoning~\cite{johnson2017inferring,jabri2016revisiting,hu2017learning}. This issue is especially severe when dealing with the more challenging image paragraph captioning task, where much more fine-grained and detailed paragraphs are expected to be generated from the given image.
Hence, it is arguably impossible to build a practical image-to-language system without machine reasoning.

An emerging line of endowing machine reasoning is to execute deep reinforcement learning (RL) in the sequence prediction task of image captioning~\cite{liu2016improved,rennie2016self,ren2017deep,zhang2017actor}. As illustrated in Figure~\ref{fig:2a}, we first frame the traditional encoder-decoder image captioning into a decision-making process, where the visual encoder can be viewed as Visual Policy (VP) that decides where to hold a gaze in the image, and the language decoder can be viewed as Language Policy (LP) that decides what the next word is. As highlighted in Figure~\ref{fig:2b}, the sequence-level RL-based framework directly injects the previously sampled word (sampling by probability distribution) to influence the next prediction. This brings the following two benefits: 1) the training supervision is delayed to the whole sequence generated. Hence, we can use non-differentiable sequence-level metrics such as CIDEr~\cite{vedantam2015cider} and SPICE~\cite{anderson2016spice}, which are more suitable than word-level cross-entropy loss for language quality evaluation; 2) it avoids the ``exposure bias''~\cite{ranzato2015sequence} by performing exploration over sequence compositions at a large scale, leading to fruitful sentences without undesirable overfitting.

However, existing RL-based framework neglects to turn VP into decision-making, \eg, the input of VP is identical in every step as shown in Figure~\ref{fig:2b}. This disrespects the nature of sequence prediction, where the historical visual actions (\eg, previously attended regions) should significantly influence the current visual policy.
One may argue that current visual attention based models would take a hidden memory vector from LP at each time step, which encodes historical cues.
However, as we will demonstrate in experiments, this strategy is not able to guide VP to concentrate on the correct regions due to that
1) the LP hidden vector is responsible to memorize linguistic context and hence lacks capacity for storing visual context;
2) it is crucial to exploit visual context to facilitate the production of fine-grained image description with complete story-line.

Motivated by the above observations, we propose a novel Context-Aware Visual Policy (CAVP) network for fine-grained image captioning. As shown in Figure~\ref{fig:2c}, CAVP allows the previous visual features, \ie, the previous output of CAVP, to serve as the visual context for the current action. Different from the conventional visual attention~\cite{xu2015show}, where the visual context is \emph{implicitly} encoded in a hidden RNN state vector from LP, our visual context is \emph{explicitly} considered in a sequence prediction process. Our motivation is in line with the cognitive evidence that the visual memory recall plays a crucial role in compositional reasoning~\cite{stanfill1986toward}.
As illustrated in Figure~\ref{fig:intuition}, for image sentence captioning, it is necessary to consider the related regions, \eg, the previously selected ``man'', when generating the composition ``man riding a horse''. For image paragraph captioning, while generating the interaction ``riding'' between ``man'' and ``horse'', we should memorize the regions within blue and red bounding boxes, which had already been concentrated in generating previous sentences. The proposed CAVP explicitly models visual context in visual policy network, leading to context-aware visual feature at each time step, which is more informative and is beneficial to fine-grained image captioning.

We decompose CAVP into four sub-policy networks, which together accomplish the visual decision-making task (cf. Figure~\ref{fig:3}), each of which is a Recurrent Neural Network (RNN) controlled by shared Long Short-Term Memory (LSTM) parameters and produces a soft visual attention map. As we will show in Section~\ref{sec:CAVP}, this CAVP design reduces the exponentially large search complexity to linear time. By reducing search complexity, it thus stabilizes the conventional Monte Carlo policy rollout. It is worth noting that CAVP and its subsequent language policy network can efficiently model higher-order compositions over time, \eg, relationships among objects mentioned in the generated sub-sequence.
Moreover, for generating a paragraph with a hierarchical structure of paragraph-sentence-word, we further develop a hierarchical CAVP network to exploit visual context at both sentence and word levels. We also design a hierarchical reward mechanism consisting of paragraph-level and sentence-level rewards.

The whole framework is trained end-to-end using an actor-critic policy gradient with a self-critic baseline~\cite{rennie2016self}.
It is worth mentioning that the proposed CAVP can be seamlessly integrated into any policy-based RL models~\cite{sutton1998reinforcement}.
We show the effectiveness of the proposed CAVP through extensive experiments on the MS-COCO image sentence captioning benchmark~\cite{lin2014microsoft} and Stanford image paragraph captioning dataset~\cite{krause2016paragraphs}.
In particular, we significantly improve every SPICE~\cite{anderson2016spice} compositional scores such as object, relation, and attribute without optimizing on it. We also show promising qualitative results of visual policy reasoning over the time of generation.

\section{Related Work}
\subsection{Image Sentence Captioning}
Inspired by the recent advances in machine translation \cite{bahdanau2014neural}, existing image captioning approaches \cite{vinyals2015show, xu2015show, lu2017knowing, anderson2017bottom, chen2017sca} typically follow an encoder-decoder framework, which can be considered as a neural machine translation task from image to text. It uses CNN-RNN architectures that encode an image as feature vectors by CNN~\cite{krizhevsky2012imagenet, he2016deep} and decode such vectors to a sentence by RNN~\cite{hochreiter1997long}.

More recently, attention mechanisms which allow dynamic feature vectors have been introduced to the encoder-decoder framework. Xu \textit{et al.}~\cite{xu2015show} incorporated \textit{soft} and \textit{hard} attention mechanisms to automatically focus on salient objects when generating corresponding words. Chen \etal ~\cite{chen2017sca} introduced channel-wise attention besides spatial attention. Lu \textit{et al.}~\cite{lu2017knowing} proposed a visual sentinel to deal with the non-visual words during captioning. Besides the spatial information comes from CNN feature maps, Anderson \textit{et al.}~\cite{anderson2017bottom} used an object detection network to propose salient image regions with an associated feature vector as bottom-up attention.
However, these captioning approaches only focus on the current time step's visual attention and neglect to consider the visual context over time, which is crucial for language compositions. Hence, we propose to incorporate historical visual attentions to current time step as visual context.

\subsection{Image Paragraph Captioning}
Describing images with a coherent paragraph is challenging.
A paragraph contains richer semantic content with longer and more descriptive descriptions. Moreover, a paragraph presents coherent and unified stories.
Krause~\etal~\cite{krause2016paragraphs} proposed a two-stage hierarchical recurrent neural network (RNN) to generate a generic paragraph for an image. The first RNN generates sentence topic vectors and decides how many sentences within the paragraph. The second RNN translates the topic vectors into a sentence.
Liang~\etal~\cite{liang2017recurrent} incorporated attention mechanism into the hierarchical RNN framework to focus on dynamic salient regions while generating corresponding sentences. They also extended the model with a Generative Adversarial Network (GAN) setting, to encourage coherence among successive sentences. They proposed a GAN-based model consisting of a paragraph generator and two discriminators for personalized image paragraph captioning.
Chatterjee~\etal~\cite{chatterjee2018diverse} explicitly introduced coherence vectors and global topic vectors to guide paragraph generation, pursuing the coherence among sentences. Moreover, they cast the model into a variational auto-encoder (VAE) framework to enhance the diversity of paragraphs.
Despite the performance of image paragraph captioning has been steadily improved, existing approaches neglect to consider visual context over time, resulting in the lack of correlation among sentences in a paragraph. Meanwhile, without reinforcement learning, they suffer from the ``exposure bias'' between training and sampling. To address these issues, we introduce a hierarchical CAVP model which can generate more coherent and descriptive paragraphs.

\subsection{Sequential Decision-Making}
Most recent captioning approaches are typically trained via maximum likelihood estimation
(MLE), resulting in the ``exposure bias''~\cite{ranzato2015sequence} between the training and testing phases. To mitigate it, reinforcement learning has been applied to image captioning, which introduces the notion of sequential decision-making.
The idea of making a series of decisions forces the agent to take into account future sequences of actions, states, and rewards.
In the case of image captioning, the state consists of visual features, preceding words and visual context, the action is choosing next word and visual representation, and the reward could be any metric of interest.

Several attempts have been made to apply sequential decision-making framework to image captioning.
For example, Ranzato~\etal~\cite{ranzato2015sequence} trained an RNN-based sequence model by policy gradient algorithm based on Monte Carlo search. The policy gradient was used to optimize a sentence-level reward.
Rennie \etal~\cite{rennie2016self} modified the classic REINFORCE algorithm~\cite{williams1992simple} with a learned baseline which obtained by greedy sampling under the current model to reduce variance of the rewards. As a result, for each sampled caption, it has a sentence level value indicating how good or bad this sentence is. It assumes that each token makes the same contribution towards the sentence.
Actor-Critic based method ~\cite{zhang2017actor} was also applied to image captioning by utilizing two networks as Actor and Critic respectively.
Ren \etal~\cite{ren2017deep} recast image captioning into decision-making framework and utilized a policy network to choose the next word and a value network to evaluate the policy.

In our work, we formulate the image captioning task into a sequence training framework where each word prediction policy is based on the action performed by the proposed CAVP. Our framework is optimized using policy gradient with a self-critic value which can directly optimize non-differentiable quality metrics of interest, such as CIDEr \cite{vedantam2015cider}.

\section{Approach}
\begin{figure*}
	\begin{center}
		\includegraphics[width=\linewidth]{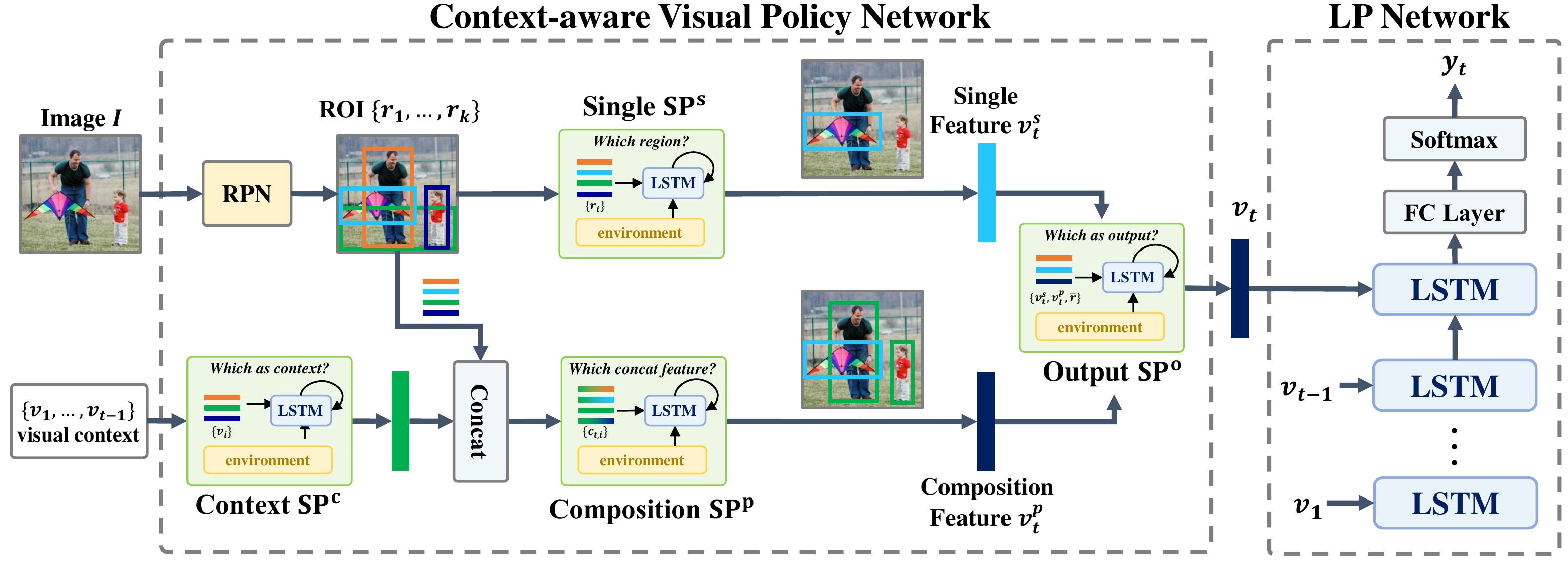}
	\end{center}
	\caption{Overview of the proposed RL-based image sentence captioning framework. It consists of the proposed CAVP for visual feature composition and the language policy for sentence generation. CAVP contains four sub-policy (SP) networks: Single SP,  Context SP, Composition SP, and Output SP. $t$ is the current time step and $y_t$ is the predicted word.}
\label{fig:3}
\end{figure*}

In this section, we elaborate the proposed fine-grained image captioning framework. We first formulate the image captioning task into a sequential decision-making process and profile the proposed models in Section~\ref{sec:problem}.
Then, we introduce the proposed Context-Aware Visual Policy network (CAVP) in Section~\ref{sec:CAVP} and language policy network (LP) in Section~\ref{sec:LP}. We discuss the sequence training strategy for the entire framework in Section~\ref{sec:train}.

\subsection{Overview}
\label{sec:problem}
We formulate the task of image captioning into a sequential decision-making process where an \textit{agent} interacts with the \textit{environment}, and then executes a series of \textit{actions}, so as to optimize the \textit{reward} return when accomplishing a \textit{goal}. Specifically, the \textit{agent} is the captioning model consisting of a context-aware visual policy network (CAVP) and a language policy network (\textsc{LP}).
The \textit{goal} is to generate a language description (sentence or paragraph) $Y$ for a given image $I$.
To accomplish the goal, at each time step $t$, the \textit{action} of CAVP is to generate a visual representation $v_t$, the \textit{action} of LP is to predict a word $y_t$.
The observed \textit{state} is the image $I$, the visual context $\{v_1, \cdots v_{t-1}\}$, and the predicted words $\{y_1, \cdots y_{t-1}\}$ so far.
The \textit{environment} is the image $I$ to be captioned.
The \textit{reward} could be any evaluation metric score between the ground-truth and the prediction.

Fig.~\ref{fig:3} illustrates the overview of the proposed image sentence captioning framework.
At each time step $t$, the CAVP takes image $I$ and visual context $\{v_1, \cdots v_{t-1}\}$ as input to produce a visual representation $v_t$. The LP takes the visual representation $v_t$ and the preceding word $y_{t-1}$ as input to predict the next word $y_t$.

\begin{figure}
\centering
	\includegraphics[width=\linewidth]{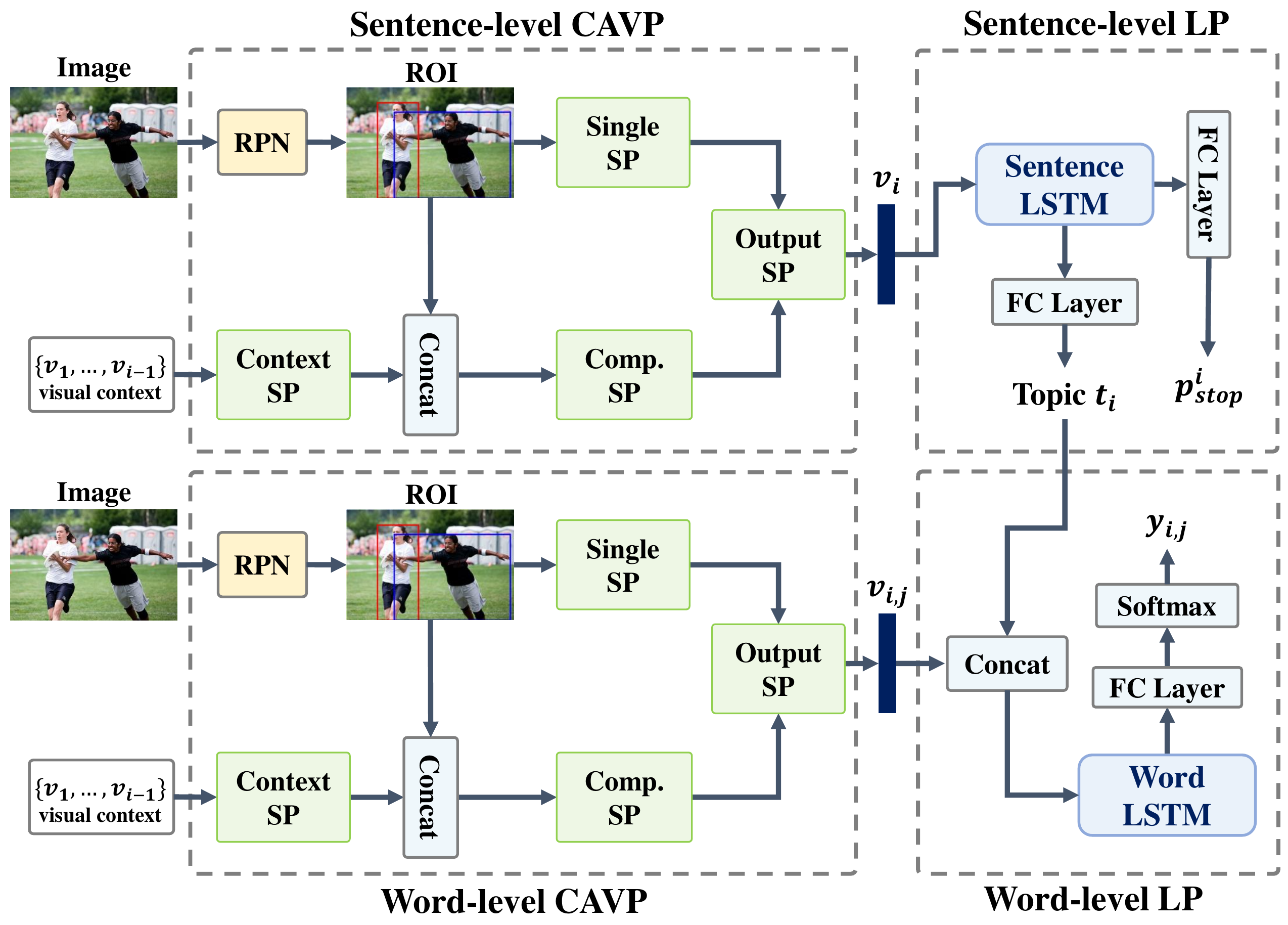}
    \caption{Overview of the proposed hierarchical CAVP-LP framework for image paragraph captioning, consisting of a sentence-level CAVP-LP and a word-level CAVP-LP.}
    \label{fig:im2p_model}
\end{figure}

Fig.~\ref{fig:im2p_model} illustrates the overview of the proposed image paragraph captioning framework.
The task of generating a paragraph could be accomplished into two steps, \textit{i.e.}, 1) producing a series of topic vectors $t$, and 2) translating each topic vector $t_i$ into a sentence of words. In particular, we design a hierarchical CAVP-LP architecture for paragraph captioning. We first utilize a sentence-level CAVP-LP which takes image $I$ and visual context $\{v_1, \cdots v_{i-1}\}$ as input and produces a visual representation $v_i$, a topic vector $t_i$ and a stop probability $p_{stop}^i$. Then, the topic vector $t_i$ is injected into a word-level LP to constraint the generation of each word $y_{i,j}$ of the $i$-th sentence.

\subsection{Context-Aware Visual Policy Network}
\label{sec:CAVP}

To generate a fine-grained description, we perform a series of complicated reasoning processes by decomposing the Context-Aware Visual Policy network (CAVP) into four sub-policy networks (SP): 1) a single SP to obtain the on-going task representation; 2) a context SP to infer task-related context; 3) a composition SP to model the relationship between the context and the on-going task; 4) an output SP to identify the efficacy of context.
We first elaborate the four sub-policy networks of CAVP in Section~\ref{sec:sp} and then introduce the hierarchical CAVP designed for paragraph captioning in Section~\ref{sec:hi_vp}.

\subsubsection{Sub-Policy Networks}
\label{sec:sp}

In general, a sub-policy network $\mathrm{SP}$ encodes the observed state by an RNN with LSTM cell~\cite{hochreiter1997long} and performs a real-valued action by soft attention selection. The selection can be considered as an approximation of Monte Carlo rollouts, reducing the sampling variance~\cite{zhang2018grounding} caused by the diverse image regions.
Specifically, at time step $t$, a sub-policy network as an \emph{agent} observes a \emph{state} $s_t$ and performs an \emph{action} $a_t \sim \pi(a_t|s_t;\theta)$. It then translates this \emph{action} into a representation $f$ as a weighted sum of a series of input features $Q_t=\{q_{1}, \cdots , q_{d}\}$, where $d$ is the number of features. Here, without loss of generality, we first introduce the general structure of the sub-policy networks denoted as SP without any superscripts. The general formulation is given by:
\begin{equation} \label{eqn:sp}
f = \mathrm{SP}(s_t, Q_t) = \sum_{i=1}^d \pi(a_t=i) q_{i}.
\end{equation}
To compute the probability distribution of the action, we follow the attention mechanism~\cite{xu2015show} as:
\begin{equation}\label{eqn:softmax}
\pi(a_t=i) = \mathrm{softmax}(w_a^T \tanh(W_h h_t + W_q q_{i})),
\end{equation}
where $\pi(a_t=i) \in [0, 1]$, $w_a$, $W_h$ and $W_q$ are trainable parameters, and $h_t$ is the LSTM hidden state calculated by:
\begin{equation}
h_{t} = \mathrm{LSTM}(s_t, h_{t-1}).
\end{equation}

In this way, if we have the state $s_t$ and the input features $Q_t$ at each time step, the sequence decision-making is known.

Next, we elaborate the implementation of each sub-policy network by introducing the corresponding state $s_t$ and the input features $Q_t$. We use a superscript to distinguish four sub-policy networks, \textit{i.e}, $\cdot^s$ for single SP, $\cdot^c$ for context SP, $\cdot^p$ for composition SP, and $\cdot^o$ for output SP.

\noindent \textbf{Single Sub-policy Network}
Before Single sub-policy network, we first use Faster R-CNN~\cite{ren2015faster} to extract image region features $\{r_1, \cdots, r_k\}$ from image $I$, where $k$ is the number of regions.
The observed state $s_t^s$ at time step $t$ consists of the previous LSTM hidden state $h_{t-1}^l$ of the language policy network, concatenated with the mean-pooled region features $\bar{r} = \frac{1}{k} \sum_{i=1}^k r_i$, and word embedding of the preceding word $y_{t-1}$:
\begin{equation}
s_t^s = [h_{t-1}^l, \bar{r}, W_e\Pi(y_{t-1})],
\end{equation}
where $W_e \in \mathbb{R}^{E\times\Sigma}$ is a word embedding matrix of a vocabulary learned from scratch, and $\Pi$ is a one-hot encoding function.
The input features at each time step are the detected region features, \ie $Q_t^s = \{r_1, r_2, \cdots, r_k\}$.
The output of the single sub-policy network is the single feature at time step $t$:
\begin{equation}
v_t^s = f_t^s = \mathrm{SP}^s(s_t^s, Q_t^s),
\end{equation}
which is in turn fed into the subsequent output SP.

\noindent \textbf{Context Sub-policy Network}
\label{sec:context}
At time step $t$, visual context includes the historical visual outputs $\{v_1, \cdots, v_{t-1}\}$. However, not every visual context is useful for the current word generation. Therefore, we introduce the context sub-policy network $\mathrm{SP}^c$ to choose the most informative context and combine it with the detected region features.
In particular, we define the observed state as:
\begin{equation}
s_t^c = [h_{t-1}^l, \bar{r}, W_e\Pi(y_{t-1})],
\end{equation}
and the input features as $Q_t^c = \{v_1, \cdots, v_{t-1}\}$.

By the context sub-policy network, we get the visual context representation $f_t^c$ at time step $t$ as Eqn.~\ref{eqn:sp}. Then we fuse $f_t^c$ with region features into context features $c_{t,i}$ as:
\begin{equation}
c_{t,i} = W_c^T[f_t^c; r_i], \  i = 1, 2, \cdots, k,
\end{equation}
where $[\cdot;\cdot]$ indicates the concatenation of vectors and $W_c^T$ projects context features to the original dimension as region features. The context features will be used in the composition SP.

To investigate the importance of visual context, we propose another way to represent context features, \textit{i.e.}, only considering preceding time step $t-1$ as visual context:
\begin{equation}
c_{t, i} = W_c^T[v_{t-1}, r_i], \  i = 1, 2, \cdots, k,
\end{equation}
We will discuss this approximation in Section~\ref{sec:ablation}.

\noindent \textbf{Composition Sub-policy Network}
\label{sec:composition}
The composition sub-policy network is similar to the single sub-policy network that takes the previous hidden state of the language policy network, the mean-pooled region features, and an embedding of the preceding word as observed state:
\begin{equation}
s_t^p = [h_{t-1}^l, \bar{r}, W_e\Pi(y_{t-1})].
\end{equation}
The input features of the composition sub-policy network are the context features from the context sub-policy network:
\begin{equation}
Q_t^p = \{c_{t,1}, c_{t,2}, \cdots, c_{t,k}\}.
\end{equation}
Then we take the output of composition sub-policy network as composition features at time step $t$:
\begin{equation}
v_t^p = f_t^p = \mathrm{SP}^p(s_t^p, Q_t^p).
\end{equation}

\noindent \textbf{Output Sub-policy Network}
\label{sec:output}
After obtaining the single and compositional features from Single SP and Composition SP, we produce the visual output $v_t$ at time step $t$ by Output SP. We define the observed state as:
\begin{equation}
s_t^o = [h_{t-1}^l, \bar{r}, W_e\Pi(y_{t-1})],
\end{equation}
and the input features as $Q_t^o = \{v_t^s, v_t^p, \bar{r}\}$.
Inspired by~\cite{lu2017knowing}, we append an extra feature $\bar{r}$ to input features for non-visual words.
We take the output of Output SP as the visual feature $v_t$:
\begin{equation}
v_t = f_t^o = \mathrm{SP}^p(s_t^o, Q_t^o).
\end{equation}
The visual feature $v_t$ will be used in language policy network at time step $t$ and also will be seen as visual context in subsequent time steps.

\noindent \textbf{Weight Sharing}
We notice that the observed state of above sub-policy networks are identical as:
\begin{equation}
s_t^c = s_t^s = s_t^p = s_t^o = [h_{t-1}^l, \bar{r}, W_e\Pi(y_{t-1})].
\end{equation}
To reduce the model complexity and computational overhead of CAVP, we share the LSTM parameters among those sub-policy networks in experiments. More ablation studies of the weight sharing will be detailed in Section~\ref{sec:ablation}.

\subsubsection{Hierarchical Context-aware Visual Policy Network}
\label{sec:hi_vp}
We design a hierarchical context-aware visual policy network, consisting of a sentence-level CAVP and a word-level CAVP, for image paragraph captioning. 
For the sake of simplicity, we denote CAVP as 
\begin{equation}
    v_t = \mathrm{CAVP}(R, S_t).
\end{equation}
At time step $t$, CAVP takes a set of region features $R$ and a sequential visual context $S_t$ as input and produces a sequential visual representation $v_t$.

\noindent \textbf{Sentence-level CAVP}
In a paragraph, each sentence should keep continuity to all the previous sentences. This requires that the model is aware of visual context.
We thus construct a sentence-level CAVP with the same structure of the CAVP for image sentence captioning in Section~\ref{sec:sp}.
Formally, given the region features $R=\{r_1, r_2, ..., r_k\}$ and visual context $\{v_1, v_2, ..., v_{i-1}\}$, we apply the four sub-policy networks including single SP, context SP, comp. SP and output SP.  The output of sentence-level CAVP will be used to guide the word-level visual policy network and feed into the sentence LSTM. Such process is given as:
\begin{equation}
    v_i = \mathrm{CAVP}(R, \{v_1, ..., v_{i-1}\}), \\ i=1,2,...,T_s
\end{equation}
where $v_i$ denotes the visual representation of $i$-th sentence, $T_s$ is the number of sentences.

\noindent \textbf{Word-level CAVP}
We construct a word-level CAVP to generate words in sentences. A single sentence in a paragraph generally describes a certain region of the image.  We set the visual context as the previous word-level visual representations:
\begin{equation}
    v_{i, j} = \mathrm{CAVP}(R, \{v_1, ..., v_{i-1}\}), \\ j=1,2,...,T_w
\end{equation}
where $v_{i,j}$ denotes the visual representation of the $j$-th word of the $i$-th sentence of a paragraph, $T_w$ is the length of sentence.
In this way, we can generate the whole paragraph by repeatedly applying the hierarchical language policy network.

\subsection{Language Policy Network}
\label{sec:LP}
We employ a language policy (LP) network towards generating a coherent image description. We first introduce the language policy network for image sentence captioning and then describe the hierarchical language policy network for image paragraph captioning.

At each time step, CAVP generates a context-aware visual representation that is most fitting to the current word. Language policy network take the visual representation and the hidden state $h_t^s$ of Single SP as input, then use them to update LSTM hidden state:
\begin{equation}
h_{t}^l = \mathrm{LSTM}([h_t^s, v_t], h_{t-1}^l).
\end{equation}
To compute the distribution over all words in vocabulary, we apply an FC layer to hidden state, and after softmax layer it outputs the probability distribution of each word, given by:
\begin{equation}\label{eqn:y_t}
\pi_l(y_t | y_{1:t-1}) = \mathrm{softmax} (W_y h^l_t + b_y),
\end{equation}
where $W_y$ and $b_y$ are learnable weights and biases. For a whole sentence, the distribution is calculated as the product of all time step's conditional distributions:
\begin{equation}\label{math_output}
\pi_l(y_{1:T})=\prod_{t=1}^T \pi_l(y_t|y_{1:t-1}).
\end{equation}

\subsubsection{Hierarchical Language Policy Network}
\label{sec:hi_lp}
We design a hierarchical language policy network for image paragraph captioning, consisting of a sentence-level LP and a word-level LP, which correspond to the sentence-level and word-level CAVPs, respectively. The sentence-level LP is fed by the visual representation $v_i$ from the sentence-level CAVP, while the word-level LP takes $v_{i,j}$ by the word-level CAVP as input. The sentence-level LP is designed to produce a \textit{topic vector} for each sentence and predict the number of sentences of a paragraph. Given a topic vector and visual representation for a sentence, the word-level LP generates each word to form the sentence.

\noindent \textbf{Sentence-level LP}
Sentence-level language policy network consists a one-layer LSTM and two FC-layers.
For each sentence in a paragraph, the LSTM receives visual representation $v_i$ and produces hidden state $h_i^{s}$. The hidden state  $h_i^{s}$ is used to generate a topic vector $t_i$ by linear projection as well as a distribution $p_{stop}^i$ over two states \{CONTINUE=0, STOP=1\} by a softmax classifier.   $p_{stop}^i$  indicates whether the current sentence is the last one in the paragraph.

\noindent \textbf{Word-level LP}
Given a topic vector from the sentence LSTM, the word LSTM is to generate the words to form the corresponding sentence. At each time step, we feed the topic vector concatenated with word embedding vector to word LSTM. The hidden state of the word LSTM is used to predict a distribution over all possible words in vocabulary by an FC-layer and a softmax classifier.

\subsection{Sequence Training}
\label{sec:train}
The sequence training process consists of two phases, including pre-training by supervised learning and fine-tuning by reinforcement learning.

For pre-training, we follow the traditional captioning training strategy and optimize the cross-entropy loss between the ground-truth and the probability distribution we produce.
Given a target ground-truth sequence $y_{1:T}^{gt}$ and a captioning model with parameters $\theta$, the objective is to minimize the cross-entropy loss:
\begin{equation}\label{eqn:xe_loss}
L_{S}(\theta) = -\sum_{t=1}^{T}\log(\pi_{l}(y_t^{gt} \mid y_{1:t-1}^{gt})).
\end{equation}
However, the ``teacher-forcing'' training strategy leads to ``exposure bias'' which means the model can hardly exposure to real sequential data beyond ground-truth dataset.

Therefore, at the fine-tuning stage, we adopt the REINFORCE algorithm~\cite{williams1992simple} to directly optimize the sequence-level metrics and address the exposure bias issue.
Specifically, we follow the self-critical method~\cite{rennie2016self}. First, we sample a greedy sequence $\hat{y}_{1:T}$ in greedy manner, \textit{i.e.}, sampling each word with the maximum probability. Then we Monte-Carlo sample another sequence $y_{1:T}^s$, \textit{i.e.}, sampling each word according to the probability distribution the model predicts.
The objective is to minimize the negative expected relative score:
\begin{equation} \label{eqn:sc_loss}
L_{R}(\theta) = -E_{y \sim \pi_l}[{r(y_{1:T}^s)-r(\hat{y}_{1:T})}],
\end{equation}
where $r(\cdot)$ could be any evaluation score metric, \textit{e.g.}, CIDEr, BLEU, or SPICE. We will discuss the influence of different metrics in Section~\ref{sec:reward}.

Note that the Eqn.~\eqref{eqn:sc_loss} is non-differentiable, we approximate the gradient by the REINFORCE algorithm as:
\begin{equation}\label{eqn:rl_gradient}
\nabla_{\theta}L_{R}(\theta) \approx -(r(y_{1:T}^s)-r(\hat{y}_{1:T}))\nabla_{\theta}\log \pi_\theta(y_{1:T}^s).
\end{equation}
While training, this gradient tends to increase the probability of each words in the sampled captions if $r(y_{1:T}^s)$ higher than $r(\hat{y}_{1:T})$, which can been seen as the relative baseline score, and vice versa.

A brute force search of all possible contextual regions requires $\mathcal{O}(2^N)$ complexity for multinomial combinations of $N$ image regions. For linear efficiency, we follow the ``divide and rule'' principle and divide the overall search process into several compositional reasoning steps by approximating the overall visual policy network as four sub-policy networks. Each sub-policy network only needs to perform specific sub-task which only requires $\mathcal{O}(N)$ complexity, \textit{e.g.}, the single SP only needs to select one region from $N$ image regions and the context SP only needs to select one historical visual output as the current visual context. As a result, the CAVP reduces the complexity exponentially.

\subsubsection{Hierarchical Sequence Training}
To train the image paragraph captioning model, we adopt the cross-entropy loss from a single sentence to a paragraph containing several sentences. Given a target ground-truth paragraph $y^{gt}_{1:T_s\times T_w}$, where $y$ has $T_s$ sentences, and each sentence contains $T_w$ words\footnote{For simplicity, we ignore the variant length of each sentence.}. Besides a word prediction loss, we also add a sentence ending prediction loss, given:
\begin{equation}
L_{S}(\theta) = -\lambda_{w}\sum_{i=1}^{T_s\times T_w} log(\pi_l(y_{i}^{gt})) - \lambda_{s} \sum_{i=1}^{T_s}log(p_{stop}^{gt}),
\label{eqn:im2p_loss}
\end{equation}
where $\lambda_{w}$ and $\lambda_{s}$ are balancing factors.

While sampling, we run the visual policy network and sentence LSTM until the stopping probability $p_{stop}>0.5$ or after maximum number of sentences.
After training by cross-entropy loss, we also use the policy gradient method to optimize the metric score directly.

\noindent \textbf{Paragraph-level Reward}
The straightforward extend method is following Eqn.~\ref{eqn:rl_gradient}, given:
\begin{equation}\label{eqn:im2p_gradient}
\begin{split}
& \nabla_{\theta}L_{R}(\theta) \approx \\
& -(r(y_{1:T_s\times T_w}^s) - r(\hat{y}_{1:T_s\times T_w})) \nabla_{\theta}\log \pi_l(y_{1:T_s\times T_w}^s)
\end{split}
\end{equation}
where $y_{1:T_s\times T_w}^s$ is a paragraph sampled according to distribution, and $\hat{y}_{1:T_s\times T_w}$ is a greedy searched paragraph description.
But in these settings, sharing one reward in a whole paragraph is insensitive, while individual rewards for each word is unstable. A trade-off is using the sentence-level reward.

\noindent \textbf{Sentence-level Reward}
Since the model generates the paragraph sentence by sentence, each sentence is based on previous sentences, besides the evaluation of NLP metrics are designed for complete strings, it can't get each sentence's reward directly.
To get the sentence-level reward, we design a sampling schedule. For example, to get the $i$-th sentence reward, we first use the previous $i-1$ ground truth sentences to guide the model, \ie~Teacher-Forcing, then sampling the next sentence according to word distribution as $y_i^s$, or greedy search the next sentence as $\hat{y}_i$. Therefore, according to Eqn.~\eqref{eqn:rl_gradient}, given:
\begin{equation}\label{eqn:im2p_gradient2}
\begin{split}
\nabla_{\theta}L_{R}(\theta) \approx
-(r(y_{1:T_w}^s) - r(\hat{y}_{1:T_w})) \nabla_{\theta}\log \pi_l(y_{1:T_w}^s)
\end{split}
\end{equation}

\subsubsection{Behavior Cloning}
The learning would be easier if we have some additional knowledge of the output policy.
While there is no any additional knowledge in the caption datasets e.g. MS-COCO, we can use a language parser \cite{toutanova2003feature} as an existing expert output policy that can be used to provide additional supervision.
More generally, if there is an expert output policy $\pi^e$ that predicts a reasonable output policy $\pi^o$, we can first pre-train our model by behavioral cloning from $\pi^e$. This can be done by minimizing the KL-divergence $D_{KL}(\pi^e || \pi^o)$ between the expert output policy $\pi^e$ and our output policy $\pi^o$, and simultaneously minimizing the captioning loss $L_{XE}$ with expert output policy $\pi^e$. This supervised behavioral cloning from the expert output policy can provide a good set of initial parameters in our output sub-policy network.
Note that the above behavioral cloning procedure is only done at cross-entropy training time to obtain a supervised initialization for our model, and the expert output policy is not used at test time.

The expert output policy is not necessarily optimal, for behavioral cloning itself is not sufficient for learning the most suitable output policy for each image.
After learning a good initialization by cloning the expert output policy, our model is further trained end-to-end with gradient $\nabla_{\theta}L_{R}(\theta)$ computed using Eqn.~\eqref{eqn:rl_gradient}, where the output policy $\pi^o$ is sampled from the output policy network in our model, and the expert output policy $\pi^e$ can be discarded.

\section{Experiments on Sentence Captioning}

In this section, we first introduce the experiment settings. Then, we go through the implementation details. Finally, we report both quantitative and qualitative evaluation results, followed by detailed ablation studies.

\subsection{Experiment Settings}
\subsubsection{Dataset}
We used the most popular benchmark \textbf{MS-COCO}~\cite{lin2014microsoft} image sentence captioning dataset, which contains 82,783 images for training and 40,504 for validation. Each image is human-annotated with 5 sentence captions. As the annotations of the official test set are not publicly available, for validating model hyperparameters and offline testing, we follow the widely used ``Karpathy'' splits~\cite{karpathy2015deep} in most prior works, containing 113,287 images for training, 5,000 for validation, and 5,000 for testing. We reported the results both on ``Karpathy'' offline split and MS-COCO online test server.

\subsubsection{Metric}
The most common metrics for caption evaluation are based on $n$-gram similarity of reference and candidate descriptions.
\textbf{BLEU}~\cite{papineni2002bleu} is defined as the geometric mean of $n$-gram precision scores, with a sentence-brevity penalty.
In \textbf{CIDEr}~\cite{vedantam2015cider}, $n$-grams in the candidate and reference sentences are weighted by term frequency-inverse document frequency weights (\ie~tf-idf). Then, the cosine similarity between them is computed.
\textbf{METEOR}~\cite{banerjee2005meteor} is defined as the harmonic mean of precision and recall of exact, stem, synonym, and paraphrase matches between sentences.
\textbf{ROUGE}~\cite{lin2004rouge} is a measures for automatic evaluation for summarization systems via F-measures.

All the above metrics are originally developed for the evaluation of text summaries or machine translations. It has been shown that there exist bias between those metrics and human judgment~\cite{anderson2016spice}. Therefore, we further evaluated our model using \textbf{SPICE}~\cite{anderson2016spice} metric, which is defined over tuples that are divided into semantically meaningful categories such as objects, relations, and attributes.

\subsection{Implementation Details}
\subsubsection{Data Pre-processing}
We performed standard minimal text pre-processing: first tokenizing on white space, second converting all words into lower case, then filtering out words that occur less than 5 times, finally resulting in a vocabulary of 10,369 words. Captions are trimmed to a maximum of 16 words for computational efficiency.
To generate a set of image region features $R$, we take the final output of the region proposed network~\cite{ren2015faster} and perform non-maximum suppression. In our implementation, we used an IoU threshold of 0.7 for region proposal non-maximum suppression, and 0.3 for object class non-maximum suppression. To select salient image regions, we simply selected the top $k=36$ features in each image for computation consider.

\subsubsection{Parameter Settings}
We set the number of hidden units of each LSTM to 1,300, the number of LSTM layers to 1, the number of hidden units in the attention mechanism we described in Eqn.~\eqref{eqn:softmax} to 1,024, and the size of word embedding to 1000.
During the supervised learning for the cross-entropy process, we use Adam optimizer~\cite{kingma2014adam} with base learning rate of 5e-4 and shrink it by 0.8 every 3 epochs. We start reinforcement learning after 37 epochs, we use Adam optimizer with base learning rate of 5e-5 and shrink it by 0.1 every 55 epochs.
We set the batch size to 100 images and train up to 100 epochs. During the inference stage, we use a beam search size of 5.
While training Faster R-CNN, we follow~\cite{anderson2017bottom} and first initialize it with ResNet-101~\cite{he2016deep} pretrained with classification on ImageNet~\cite{russakovsky2015imagenet}, then fine-tune it on Visual Genome~\cite{krishna2017visual} with attribute labels.

\subsection{Comparisons to State-of-The-Arts}
\subsubsection{Comparing Methods}
\noindent \textbf{Traditional Approaches}
We first compared our models to classic methods including~\textbf{Google NIC}~\cite{vinyals2015show}, \textbf{Hard Attention}~\cite{xu2015show}, \textbf{Adaptive Attention}~\cite{lu2017knowing} and \textbf{LSTM-A}~\cite{yao2016boosting}. These methods follow the popular encoder-decoder architecture, trained with cross-entropy loss between the predicted and ground-truth words, that is, no sequence training is applied.

\noindent \textbf{RL-based Approaches}
We also compared our models to the RL-based methods including \textbf{PG-SPIDEr-TAG}~\cite{liu2016improved}, \textbf{SCST}~\cite{rennie2016self}, \textbf{Embedding-Reward}\cite{ren2017deep}, and \textbf{Actor-Critic}~\cite{zhang2017actor}. These methods use sequence training with various reward returns.

\subsubsection{Quantitative Analysis}
\begin{table}
	\centering
	\begin{tabular}{l|ccccc}
		\hline
		Model          & B@4  & M    & R    & C     & S    \\ \hline
		Google NIC\cite{vinyals2015show}   & 32.1 & 25.7 & -    & 99.8  & -    \\
		Hard-Attention\cite{xu2015show} & 24.3 & 23.9 & -    & -     & -    \\
		Adaptive\cite{lu2017knowing}       & 33.2 & 26.6 & 54.9 & 108.5 & 19.4 \\
		LSTM-A\cite{yao2016boosting}         & 32.5 & 25.1 & 53.8 & 98.6  & -    \\ \hline
		PG-SPIDEr\cite{liu2016improved}      & 32.2 & 25.1 & 54.4 & 100.0 & -    \\
		Actor-Critic\cite{zhang2017actor}  & 34.4 & 26.7 & 55.8 & 116.2 & - \\
        EmbeddingReward\cite{ren2017deep}           & 30.4 & 25.1 & 52.5 & 93.7 & -    \\
		SCST\cite{rennie2016self}           & 35.4 & 27.1 & 56.6 & 117.5 & -    \\
		StackCap\cite{gu2017stack}       & 36.1 & 27.4 & 56.9 & 120.4 & 20.9 \\
		Up-Down\cite{anderson2017bottom}        & 36.3 & 27.7 & 56.9 & 120.1 & 21.4 \\ \hline
		Ours           & \textbf{38.6} & \textbf{28.3} & \textbf{58.5} & \textbf{126.3} & \textbf{21.6} \\ \hline
	\end{tabular}
	\caption{
		Performance comparisons on MS-COCO ``Karpathy'' offline split. B@n is short for BLEU-n, M is short for METEOR, R is short for ROUGE, C is short for CIDEr, and S is short for SPICE.
	}
    \label{tab:offline}
\end{table}
\begin{table*}[!t]
    \begin{center}\small
        \setlength{\tabcolsep}{.35em}
        \begin{tabular}{lccccccccccccccccccccc}
            \midrule
            & \multicolumn{2}{c}{BLEU-1} &  & \multicolumn{2}{c}{BLEU-2} &  & \multicolumn{2}{c}{BLEU-3} &  & \multicolumn{2}{c}{BLEU-4} &  & \multicolumn{2}{c}{METEOR} &  & \multicolumn{2}{c}{ROUGE-L} &  & \multicolumn{2}{c}{CIDEr} \\
            \cmidrule{2-3}\cmidrule{5-6}\cmidrule{8-9}\cmidrule{11-12}\cmidrule{14-15}\cmidrule{17-18}\cmidrule{20-21}
            & c5           & c40         &  & c5           & c40         &  & c5           & c40         &  & c5           & c40         &  & c5           & c40         &  & c5           & c40         &  & c5           & c40        \\
            \midrule
            Google NIC\cite{vinyals2015show} &71.3&89.5& &54.2&80.2& &40.7&69.4& &30.9&58.7& &25.4&34.6& &53.0&68.2& &94.3&94.6\\
            MSR Captivator\cite{fang2015captions}&71.5&90.7& &54.3&81.9& &40.7&71.0& &30.8&60.1& &24.8&33.9& &52.6&68.0& &93.1&93.7\\
            M-RNN\cite{mao2014deep}&71.6&89.0& &54.5&79.8& &40.4&68.7& &29.9&57.5& &24.2&32.5& &52.1&66.6& &91.7&93.5\\
            Hard-Attention\cite{xu2015show} &70.5&88.1& &52.8&77.9& &38.3&65.8& &27.7&53.7& &24.1&32.2& &51.6&65.4& &86.5&89.3\\
        Adaptive\cite{lu2017knowing} &74.8&92.0& &58.4&84.5& &44.4&74.4& &33.6&63.7& &26.4&35.9& &55.0&70.5& &104.2&105.9\\
            PG-SPIDEr-TAG\cite{liu2016improved}&75.1&91.6& &59.1&84.2& &44.5&73.8& &33.6&63.7& &25.5&33.9& &55.1&69.4& &104.2&107.1\\
            SCST:Att2all\cite{rennie2016self}&78.1&93.7& &61.9&86.0& &47.0&75.9& &35.2&64.5& &27.0&35.5& &56.3&70.7& &114.7&116.7\\
            LSTM-A$_3$ \cite{yao2016boosting}&78.7&93.7& &62.7&86.7& &47.6&76.5& &35.6&65.2& &27&35.4& &56.4&70.5& &116.0&118.0\\
            Stack-Cap\cite{gu2017stack}&77.8&93.2& &61.6&86.1& &46.8&76.0& &34.9&64.6& &27.0&35.6& &56.2&70.6& &114.8&118.3\\
            Up-Down\cite{anderson2017bottom}& \textbf{80.2} & \textbf{95.2} &  & 64.1 & 88.8 &  & 49.1 & 79.4 &  & 36.9 & 68.5 &  & 27.6 & 36.7 &  & 57.1 & 72.4 &  & 117.9 & 120.5\\
            \midrule
            Ours & 80.1 & 94.9 &  & \textbf{64.7} & \textbf{88.8} &  & \textbf{50.0} & \textbf{79.7} &  & \textbf{37.9} & \textbf{69.0} &  & \textbf{28.1} & \textbf{37.0} &  & \textbf{58.2} & \textbf{73.1} &  & \textbf{121.6} & \textbf{123.8}\\
            \midrule
        \end{tabular}
    \end{center}
    \caption{Highest ranking published image captioning results on the online MSCOCO test server. Except for BLUE-1 which is of little interest, our single model optimized with CIDEr, outperforms previously published works using all the other metrics.}
    \vspace{-3mm}
    \label{tab:online}
\end{table*}
\begin{figure}
\centering
\includegraphics[width=0.9\linewidth]{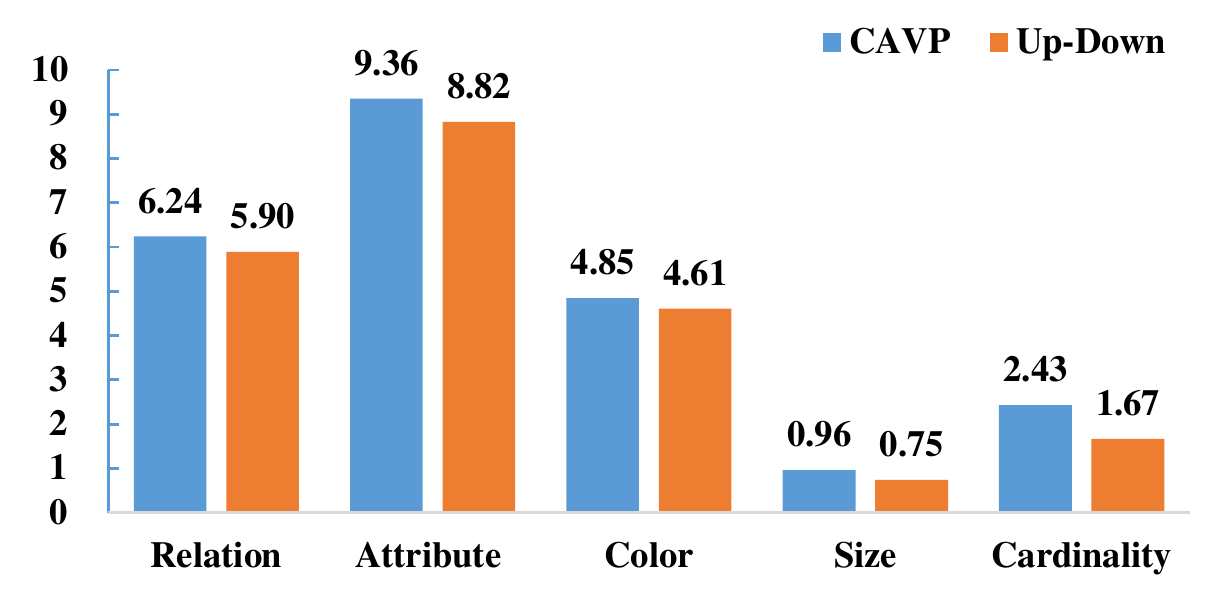}
\caption{The performance comparison of the CAVP model and the Up-Down method. All SPICE category scores are improved by CAVP.}
\label{fig:spice}
\end{figure}

As shown in Table \ref{tab:offline}, we evaluated our model compared to multiple state-of-the-art methods.
We found that almost all RL-based methods outperform traditional ones.
The reason is that RL addresses the loss-evaluation mismatch problem and included the inference process in training to address the exposure bias problem.
We can also find that our CAVP outperforms other non-context methods.
This is because the visual context information is useful for current word generation and the policy makes better decisions. In particular, we achieved state-of-the-art performance under all metrics on ``Karpathy'' test split. Table~\ref{tab:online} reports the performance comparison without any ensemble on the official MS-COCO evaluation server\footnote[1]{\url{https://competitions.codalab.org/competitions/3221\#results}}. It is worthy to note that our approach is a single captioning model while the others are based on the ensemble of multiple captioning models.

To evaluate the compositional reasoning ability of our CAVP model, we also provide SPICE semantic category scores in Fig.~\ref{fig:spice}.
Since SPICE parses the language into scene graph and compares the graph similarity, it can provide finer-grained information such as relation, attribute, color, size, and cardinally.
We can find that our CAVP model improves all SPICE semantic category scores while comparing with Up-Down\cite{anderson2017bottom} model which neglects visual context.
Specifically, the Relation score indicates the reasoning ability of object relationships, \textit{e.g.}, ``man riding horse''. The Attribute, Color, and Size scores indicate the reasoning ability of visual comparisons, \textit{e.g.}, ``small(er) cat''. Note that in most cases, the visual comparisons are implicit, for example, when we describe a cat is ``small'', it means the cat is relatively ``smaller'' than other objects.

\subsubsection{Qualitative Analysis}
\begin{figure}
	\centering
	\begin{subfigure}{0.4\linewidth}
    	\centering
		\includegraphics[height=\linewidth]{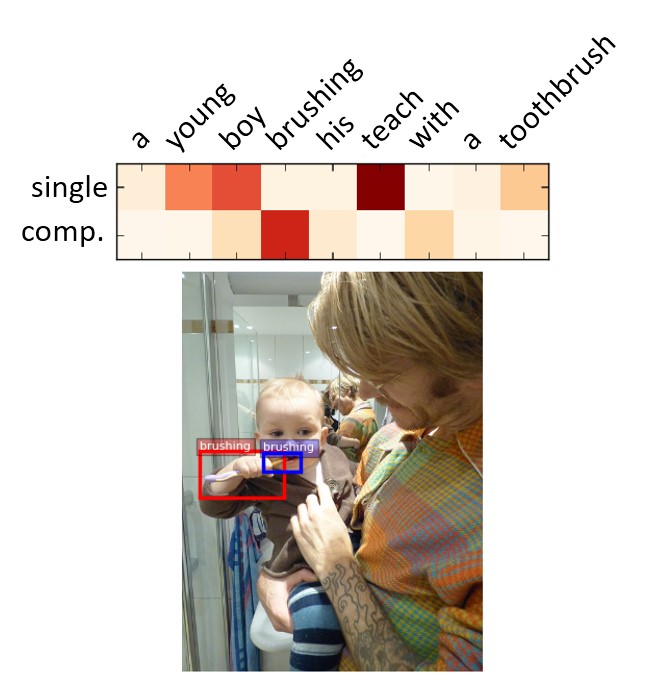}
		\caption{brushing}
        \label{fig:word_brushing}
	\end{subfigure}%
	\begin{subfigure}{0.4\linewidth}
        \centering
		\includegraphics[height=\linewidth]{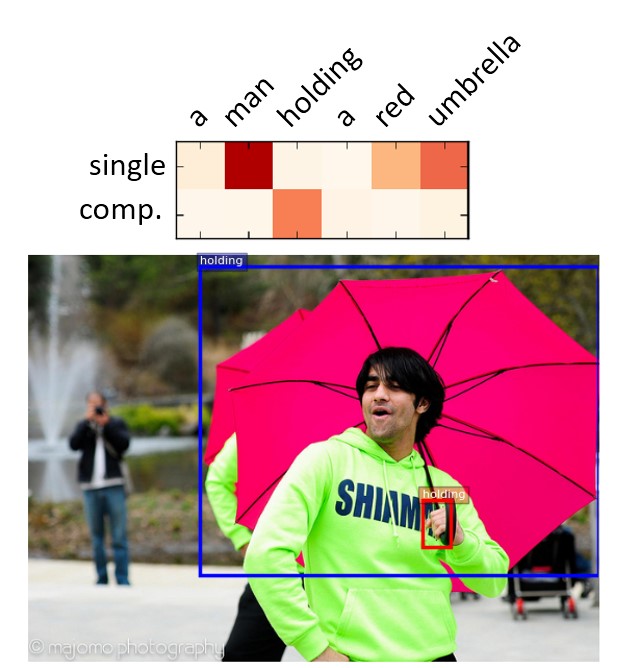}
		\caption{holding}
        \label{fig:word_holding}
	\end{subfigure}

	\begin{subfigure}{0.4\linewidth}
		\centering
		\includegraphics[height=\linewidth]{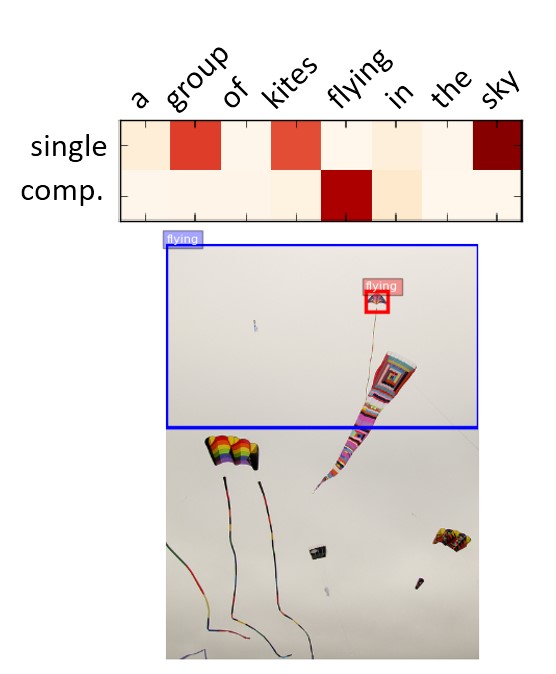}
		\caption{flying}
        \label{fig:word_flying}
	\end{subfigure}%
	\begin{subfigure}{0.4\linewidth}
		\centering
		\includegraphics[height=\linewidth]{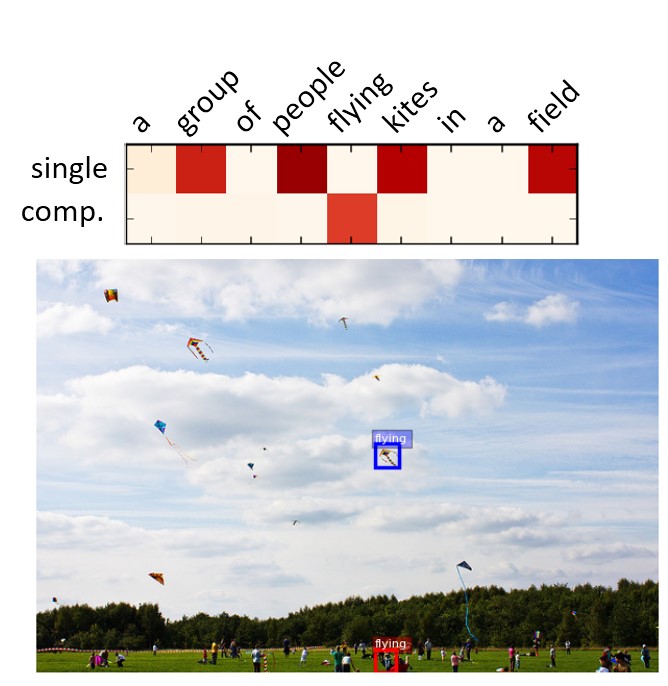}
		\caption{flying}
        \label{fig:word_flying2}
	\end{subfigure}
	\caption{
    Qualitative examples where top matrix shows the output policy network action probabilities and the bottom image shows the decision with maximum probability for composition features. The red bounding boxes are the context regions and the blue bounding boxes are the current regions which concatenated with context regions.
}
\vspace{-4mm}
	\label{fig:word}
\end{figure}

To better reveal our CAVP model, we show some qualitative visualizations as well as the output of sub-policy network's predictions in Figure ~\ref{fig:word}.
Take Figure~\ref{fig:word_brushing} as an example, after we generated ``a young boy'', we first focus on the visual context, \textit{i.e.}, the boy's hand which is holding something. Then we want to find visual regions that the boy is holding, so we focus on the toothbrush in the boy's mouth. Finally, focusing on both hand and toothbrush, we generated the exact word ``brushing''. In Figure ~\ref{fig:word_flying} and ~\ref{fig:word_flying2}, although the model generated the same words, the context of the two words are different.
In Figure~\ref{fig:word_flying}, the context is ``kite'' for captioning the relation between the kite and sky, while in Figure~\ref{fig:word_flying2}, the context is ``people'' for captioning the action of the people.
By applying the CAVP model, we can generate those captions both successfully with deep understanding of image scenes.

\begin{figure*}
	\begin{subfigure}{.48\linewidth}
	\centering
	\includegraphics[width=\linewidth]{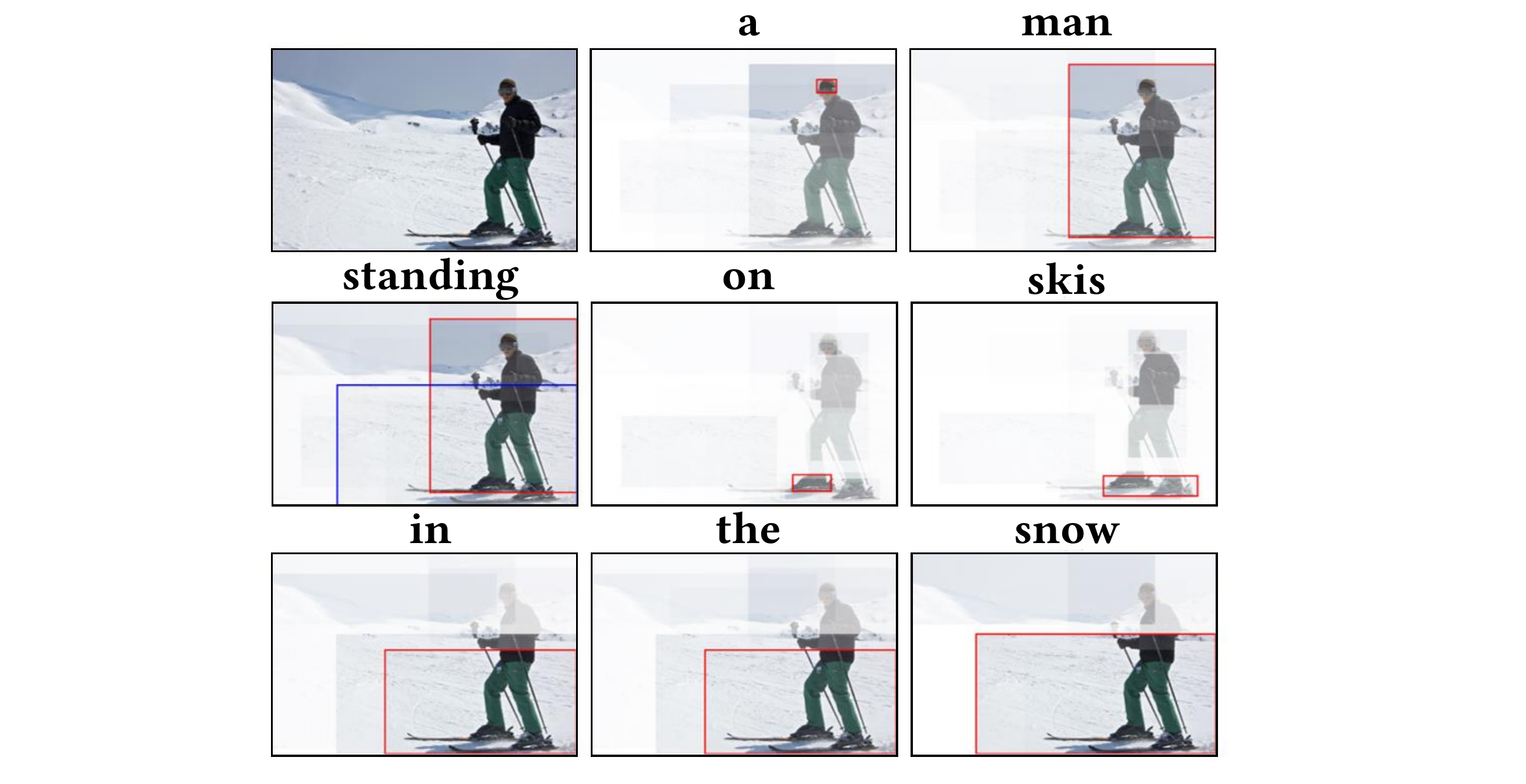}
	\caption{a man standing on skis in the snow}
    \end{subfigure}
\begin{subfigure}{.48\linewidth}
	\centering
	\includegraphics[width=\linewidth]{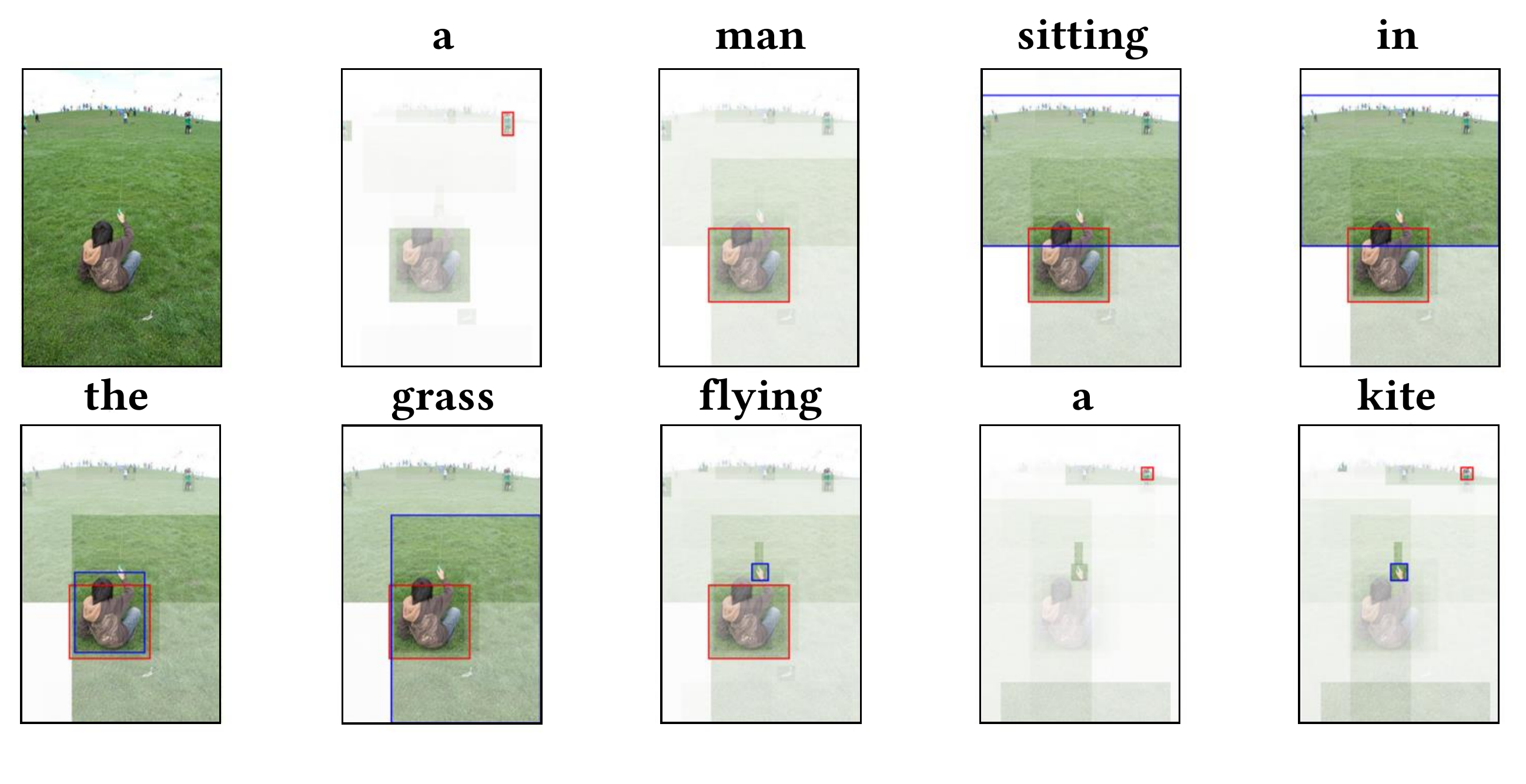}
	\caption{a man sitting in the grass flying a kite}
\end{subfigure}
\caption{For each generated word, we visualized the attended image regions, outlining the region with the maximum policy probability in bounding box. The red bounding boxes are the visual context representation regions and the blue bounding boxes are the regions decided by single policy network.}
\label{fig:sent}
\end{figure*}

Besides showing a single important word of the generated sequence, we also visualize the whole policy decision across the whole sentence generation in Figure~\ref{fig:sent}.
Take the first sentence as an example, we notice that our context-aware model can not only focus on some single objects such as ``man'', ``skis'', and ``snow'', but also the compositional word ``standing'', connecting ``man standing in snow''.

\subsection{Ablation Studies}
We extensively investigated ablation structures and settings of the CAVP model to gain insights into how and why it works.
\subsubsection{Architecture}
\label{sec:ablation}
We investigate multiple variants of the CAVP model.
\begin{itemize}[leftmargin=.1in]
\item
\textbf{Up-Down~\cite{anderson2017bottom}}: The CAVP degrades to the existing Up-Down model if we only use the single sub-policy network.
\item
\textbf{CAVP\_scratch}: In CAVP, the context sub-policy network only takes the last visual feature as visual context and the sub-policy network is trained from scratch rather than using expert policy.
\item
\textbf{CAVP\_cloning}: The context sub-policy network takes the last visual feature as visual context. The output sub-policy network is behavior cloned from expert policy.
\item
\textbf{CAVP\_non-sharing}: The context sub-policy does not share weights with the other three sub-policy networks.
\end{itemize}

\begin{table}
\centering
\begin{tabular}{cl|ccccc}
\hline
& Model & B@4 & M & R & C & S \\ \hline
1 & \textbf{Up-Down~\cite{anderson2017bottom}} & 37.5 & 27.7 & 57.9 & 121.9 & 21.0 \\
2 & \textbf{CAVP\_scratch} & 37.8 & 28.0 & 58.2 & 124.5 & 21.3 \\
3 & \textbf{CAVP\_cloning} & 38.3 & 27.8 & 58.0 & 124.6 & 21.4 \\
\hline
4 & \textbf{CAVP\_non-sharing} & 38.3 & 28.2 & 58.4 & \textbf{126.4} & 21.6 \\
5 & \textbf{CAVP} & \textbf{38.6} & \textbf{28.3} & \textbf{58.5} & 126.3 & \textbf{21.6} \\
\hline
\end{tabular}
\caption{Ablation performance on MS-COCO. B@n is short for BLEU-n, M is short for METEOR, R is short for ROUGE, C is short for CIDEr, and S is short for SPICE.}
\label{tab:ablation}
\end{table}
\begin{table}
\centering
\begin{tabular}{l|ccc}
\hline
Model & \# of Parameters & Training & Testing \\
\hline
\textbf{Up-Down~\cite{anderson2017bottom}} & 77.6M & 66 & 44.40\\
\textbf{CAVP\_non-sharing} & 108.5M & 78 & 58.51\\
\textbf{CAVP} & 83.0M & 72 & 56.48\\
\hline
\end{tabular}
\caption{Efficiency comparison in terms of parameter number, training time (hour) and testing time (ms/image). Experiments are performed on two Nvidia 1080Ti GPUs.
}
\label{tab:complexity}
\end{table}
Table~\ref{tab:ablation} reports the performance comparison between the CAVP model and its variants on MS-COCO dataset. We can have the following observations: (a) The performance improvements of the other four models over Up-Down method~\cite{anderson2017bottom} indicates the effectiveness of visual context for fine-grained image captioning. (b) The \textbf{CAVP\_scratch} and \textbf{CAVP\_cloning} obtain comparable performance. This shows that the off-the-shelf language parser is not very suitable to the visual-language task and the output sub-policy network can be learned from scratch without any expert policy guiding. (c)  \textbf{CAVP} outperforms  \textbf{CAVP\_scratch} and \textbf{CAVP\_cloning}. By memorizing historical visual context rather than only using the last visual feature, \textbf{CAVP} is able to generate more effective visual representations for subsequent sentence/paragraph generation. 

Table 4 reports the parameter number, training and testing time costs. From the results, we can see that the \textbf{CAVP} model slightly increase the parameter number, training and testing computational overhead as compared to the existing up-down method~\cite{anderson2017bottom}. Moreover, by sharing parameters among the four sub-policy networks, \textbf{CAVP} has fewer parameters and lower computational cost than the model without parameter sharing.

\subsubsection{Reward}
\label{sec:reward}
\begin{table}
\centering
\begin{tabular}{c|ccccc}
\hline
\multirow{2}{*}{\begin{tabular}[c]{@{}c@{}}Training\\ Metric\end{tabular}} & \multicolumn{5}{c}{Evaluation Metric} \\
 & BLEU4 & ROUGE & METEOR & CIDEr & SPICE \\ \hline
BLEU & \textbf{38.8} & 57.7 & 27.3 & 114.5 & 20.7 \\
ROUGE & 38.1 & \textbf{59.1} & 27.8 & 120.0 & 20.8 \\
METEOR & 33.6 & 57.6 & \textbf{29.6} & 113.0 & 22.8 \\
CIDEr & 38.3 & 58.4 & 28.2 & \textbf{126.4} & 21.6 \\
SPIDEr & 37.8 & 58.0 & 27.8 & 125.3 & \textbf{23.1} \\ \hline
\end{tabular}
\caption{
Ablation performance on the MS-COCO ``Karpathy'' offline split with respect to various metrics as the reward.
}
\label{tab:reward}
\end{table}
For sequence training by policy gradient, the reward function $r(\cdot)$ can be any metrics, \eg~BLEU, ROUGE, METEOR, CIDEr, and SPIDEr~\cite{liu2016improved} (which combining the CIDEr and SPICE scores equally as the reward). Optimizing for different metrics leads to different performance. In general, as shown in Table~\ref{tab:reward}, we found that optimizing for a specific metric results in the best performance on the same metric. And optimizing for CIDEr and SPIDEr gives the best overall performance, but the SPIDEr is more time consuming as the SPICE metric evaluation is very slow. Thus, we chose the CIDEr as the optimizing objective in most of our experiments.

\section{Experiments on Paragraph Captioning}
\begin{table*}
\centering
\begin{tabular}{l|cccccc}
  \hline
  & METEOR & CIDEr & BLEU-1 & BLEU-2 & BLEU-3 & BLEU-4 \\
  \hline
  Sentence-Concat~\cite{krause2016paragraphs} & 12.05 & 6.82 &  31.11 & 15.10 & 7.56 & 3.98\\
  Template~\cite{krause2016paragraphs} & 14.31 & 12.15 &  37.47 & 21.02 & 12.30 & 7.38\\
  DenseCap-Concat~\cite{krause2016paragraphs} & 12.66 & 12.51 &  33.18 & 16.92 & 8.54 & 4.54\\
  Image-Flat~\cite{krause2016paragraphs} & 12.82 & 11.06 &  34.04 & 19.95 & 12.20 & 7.71 \\
  \hline
  Regions-Scratch~\cite{krause2016paragraphs} & 13.54 & 11.14 & 37.30 & 21.70 & 13.07 & 8.07\\
  Regions-Hierarchical~\cite{krause2016paragraphs} & 15.95 & 13.52 & 41.90 & 24.11 & 14.23 & 8.69 \\
  RTT-GAN~\cite{liang2017recurrent} & 17.12 & 16.87 & 41.99 & 24.86 & 14.89 & 9.03 \\
  RTT-GAN*~\cite{liang2017recurrent} & \textbf{18.39} & 20.36 & \textbf{42.06} & 25.35 & 14.92 & 9.21 \\
  \hline
  Hierarchical CAVP\_CIDEr & 16.79 & 20.94 & 41.38 & 25.40 & 14.93 & 9.00 \\
  Hierarchical CAVP\_BLEU & 16.83 & \textbf{21.12} & 42.01 & \textbf{25.86} & \textbf{15.33} & \textbf{9.26} \\
  \hline
  Human & 19.22 & 28.55 & 42.88 & 25.68 & 15.55 & 9.66\\
  \hline
\end{tabular}
\caption{Performance comparison on image paragraph captioning task. The proposed models outperform the state-of-the-art methods in terms of most metrics.
\vspace{-2mm}}
  \label{tab:im2p_res}
\end{table*}

\begin{table*}
\centering
\begin{tabular}{l|cccccc}
  \hline
  & METEOR & CIDEr & BLEU-1 & BLEU-2 & BLEU-3 & BLEU-4 \\
  \hline
  Single Policy & 16.29 & 16.36 & 40.11 & 22.31 & 12.39 & 6.86\\
  Sentence CAVP & 16.41 & 17.29 & 41.79 & 24.47 & 13.67 & 7.82\\
  Hierarchical CAVP\_XE & \textbf{17.14} & 19.63 & \textbf{42.49} & 25.80 & 15.04 & 9.00 \\
  \hline
  Hierarchical CAVP\_CIDEr & 16.79 & 20.94 & 41.38 & 25.40 & 14.93 & 9.00 \\
  Hierarchical CAVP\_BLEU & 16.83 & \textbf{21.12} & 42.01 & \textbf{25.86} & \textbf{15.33} & \textbf{9.26} \\
  \hline
\end{tabular}
\caption{Ablation performance on image paragraph captioning task.
\vspace{-2mm}}
  \label{tab:im2p_comp}
\end{table*}

\subsection{Experiment Settings}
We conducted the experiments on the publicly available Stanford image-paragraph dataset collected by Krause~\etal~\cite{krause2016paragraphs}, which is divided into three subsets, including 14,575 images for training, 2,487 for validation and 2,489 for testing. Each image is annotated with one paragraph that contains an average of 5.7 sentences. where each sentence contains 11.9 words in average. For performance evaluation, we reported six widely used performance metrics: BLEU-\{1,2,3,4\}, METEOR, and CIDEr.

We performed the standard minimal textual pre-processing as in Section 4.2.1, leading to a vocabulary of 4,237 words. To generate a set of image region features, we followed the dense captioning~\cite{densecap} settings. In particular, we first resized each image so that its longest edge is 720 pixels and passed it through VGG-16~\cite{simonyan2014very} network. Then, we extracted 50 region features in 4,096 dimensions.
For policy network, we set LSTM size to 512, the number of hidden units in Eqn.~\eqref{eqn:softmax} to 512, and embedding dimension to 512. We set $\lambda_{w}=1.0$ and $\lambda_{s}=5.0$ in Eqn.~\eqref{eqn:im2p_loss}. During the training, we used Adam optimizer~\cite{kingma2014adam} with base learning rate of 5e-4 and shrank it by 0.8 every 20 epochs. We set the batch size to 64 images and trained up to 75 epochs for cross-entropy loss and up to 150 epochs for RL loss. Besides, we set maximum number of sentences to 6 and maximum sentence length to 30 words.

\begin{figure}
	\centering
	\includegraphics[width=\linewidth]{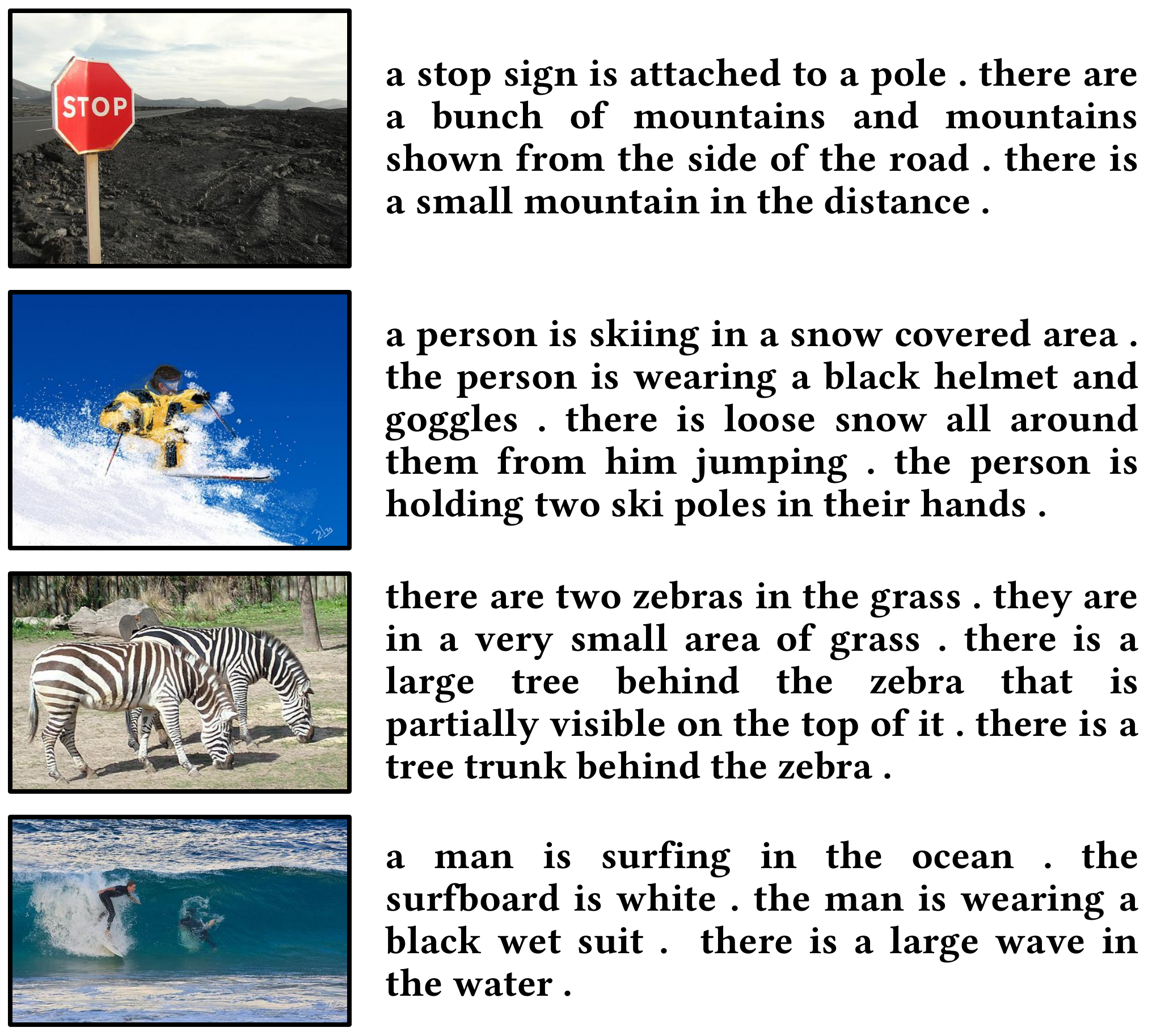}
\caption{Examples of image paragraph captioning results of our model. For each image, a paragraph description with a variable number of sentences is generated.}
\label{fig:im2p_examples}
\end{figure}

\subsection{Comparison to State-of-the-Arts}
We compared our hierarchical CAVP model to the following state-of-the-art methods.
\textbf{Sentence-Concat}~\cite{krause2016paragraphs} combines five sentences sampled from a sentence captioning model trained on MS-COCO dataset. \textbf{Image-Flat}~\cite{krause2016paragraphs} directly treats a paragraph as a long sentence and applies a standard image captioning method~\cite{karpathy2015deep}. \textbf{Template}~\cite{krause2016paragraphs} converts a structured representation of images into text via a pre-defined template. \textbf{DenseCap-Concat}~\cite{krause2016paragraphs} concatenates DenseCap~\cite{densecap} predictions to form a paragraph. \textbf{Region-Scratch}~\cite{krause2016paragraphs} uses a flat model, which initialized from scratch, to decode paragraph. \textbf{Region-Hierarchical}~\cite{krause2016paragraphs} uses a hierarchical structure contained a sentence RNN and a word RNN.
\textbf{RTT-GAN}~\cite{liang2017recurrent} is an recurrent topic-transition generative adversarial network coupled with an attention mechanism proposed recently. \textbf{RTT-GAN*}~~\cite{liang2017recurrent} is the version using additional training data.
Moreover, we performed a \textbf{Human} evaluation by collecting an additional paragraph for 500 randomly chosen images.

Table~\ref{tab:im2p_res} reports the performance comparison of image paragraph captioning on the Stanford image-paragraph dataset.
We found that the proposed Hierarchical CAVP model optimized with either CIDEr or BLEU both outperforms the state-of-the-art methods in terms of most metrics.
Note that even comparing with \textbf{RTT-GAN*}~\cite{liang2017recurrent} which uses additional training data, the proposed model achieves better performance in terms of most metrics. Moreover, \textbf{Human} produce superior description to all the automatic methods, especially in CIDEr and METEOR which are more correlated with human judgment.

Figure~\ref{fig:im2p_examples} presents some examples generated by the proposed models. We can find that our models can generate successive sentences with a storyline. For example, the paragraph for the image in the first row moves its attention from near to far. The successive sentences first focus on the nearest sign, then mountains at the side of road, and the farthest mountain in the background finally.

\subsection{Ablation Studies}
We conducted ablation experiments to compare the proposed model and its following variants. \textbf{Single Policy} and \textbf{Sentence CAVP} treat a paragraph as a long sentence. While \textbf{Single Policy} only uses the sentence-level single sub-policy network without any visual context, \textbf{Sentence CAVP} uses the sentence-level CAVP without hierarchical fortification. \textbf{Hierarchical CAVP\_XE} is the proposed hierarchical CAVP trained by cross-entropy loss.

Table~\ref{tab:im2p_comp} reports the performance comparison among the proposed modes and the variants. From the results, we can obtain the following observations. (a) The \textbf{Sentence CAVP} outperforms \textbf{Single Policy} in terms of all the metrics. This indicates that the context-aware visual policy network can generate better long sentences by exploiting visual context. (b) \textbf{Hierarchical CAVP\_XE} performs better than \textbf{Sentence CAVP} by using sentence-level and word-level visual policies augmented with visual context at both levels. (c) \textbf{Hierarchical CAVP\_CIDEr} or \textbf{Hierarchical CAVP\_BLEU} achieves performance improvements in terms of some metrics and causes performance degradation on the others as compared to \textbf{Hierarchical CAVP\_XE}. The main reason is the lack of sufficient ground-truth paragraphs for model training. There is only one ground-truth paragraph for each image in the dataset. Given more ground-truth paragraphs, the models optimized by CIDEr or BLEU would be more superior over that by cross entropy, as shown in the evaluation of image sentence captioning, where each image has five ground-truth captions. (d) \textbf{Hierarchical CAVP\_BLEU} performs better than \textbf{Hierarchical CAVP\_CIDEr}. This indicates that BLEU is more stable than CIDEr when dealing with limited ground-truth and small dataset.

\section{Conclusion}
In this paper, we proposed a novel Context-Aware Visual Policy network (CAVP) for fine-grained image-to-language generation, including both image sentence captioning and image paragraph captioning. Superior to existing RL-based methods, the proposed CAVP based framework takes advantage of visual context in compositional visual reasoning, which is beneficial for image captioning. Compared against traditional visual attention which only fixes a single image region at every step, CAVP can attend to complex visual compositions over time. To the best of our knowledge, CAVP is the first RL-based image captioning model which incorporates visual context into sequential visual reasoning. We conducted extensive experiments as well as ablation studies to investigate the effectiveness of CAVP. The experimental results have shown that the proposed approach can significantly boost the performances of the RL-based image captioning methods and achieves top ranking performances on MS-COCO server and Stanford image paragraph captioning dataset. We will continue our future works in two directions. First, we will integrate the visual policy and language policy into a Monte Carlo search strategy for image sentence/paragraph captioning. Second, we will also apply CAVP to other sequential decision-making tasks such as visual question answering and visual dialog.
\ifCLASSOPTIONcompsoc
  \section*{Acknowledgments}
\else
  \section*{Acknowledgment}
\fi

This work was supported by the National Natural Science Foundation of China (NSFC) under Grants 61622211, 61620106009 and 61525206 as well as the Fundamental Research Funds for the Central Universities under Grant WK2100100030.

\ifCLASSOPTIONcaptionsoff
  \newpage
\fi

\bibliography{citation}
\bibliographystyle{IEEEtran}

\begin{IEEEbiography}[{\includegraphics[width=1in,height=1.25in,clip,keepaspectratio]{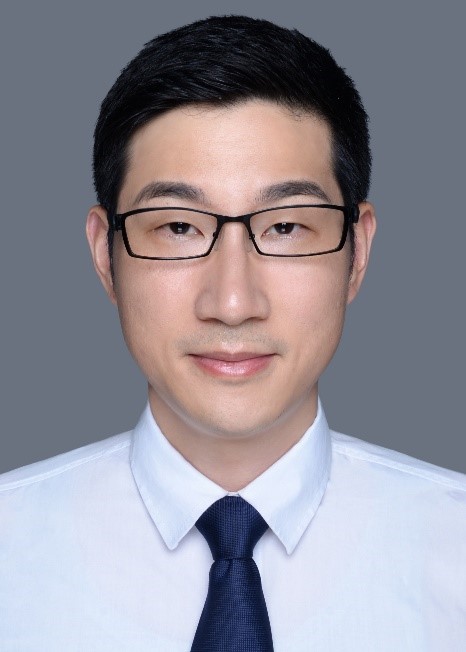}}]
{Zheng-Jun Zha} (M’08) received the B.E. and Ph.D. degrees from the University of Science and Technology of China, Hefei, China, in 2004 and 2009, respectively. He is currently a Full Professor with the School of Information Science and Technology, University of Science and Technology of China, the Vice Director of National Engineering Laboratory for Brain-Inspired Intelligence Technology and Application. He was a Researcher with the Hefei Institutes of Physical Science, Chinese Academy of Sciences, from 2013 to 2015, a Senior Research Fellow with the School of Computing, National University of Singapore (NUS), from 2011 to 2013, and a Research Fellow there from 2009 to 2010. His research interests include multimedia analysis, retrieval and applications, as well as computer vision etc. He has authored or coauthored more than 100 papers in these areas with a series of publications on top journals and conferences. He was the recipient of multiple paper awards from prestigious multimedia conferences, including the Best Paper Award and Best Student Paper Award in ACM Multimedia, etc. He serves as an Associated Editor of IEEE Trans. on Circuits and Systems for Video Technology.
\end{IEEEbiography}

\begin{IEEEbiography}[{\includegraphics[width=1in,height=1.25in,clip,keepaspectratio]{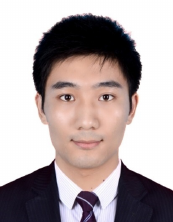}}]
{Daqing Liu} received the B.E. degree in Automation from Chang'an University,
Xi'an, China, in 2016, and currently working
toward the Ph.D. degree from the Department of Automation, University of Science and Technology of
China, Hefei, China. His research interests mainly include computer vision and multimedia.
\end{IEEEbiography}

\begin{IEEEbiography}[{\includegraphics[width=1in,height=1.25in,clip,keepaspectratio]{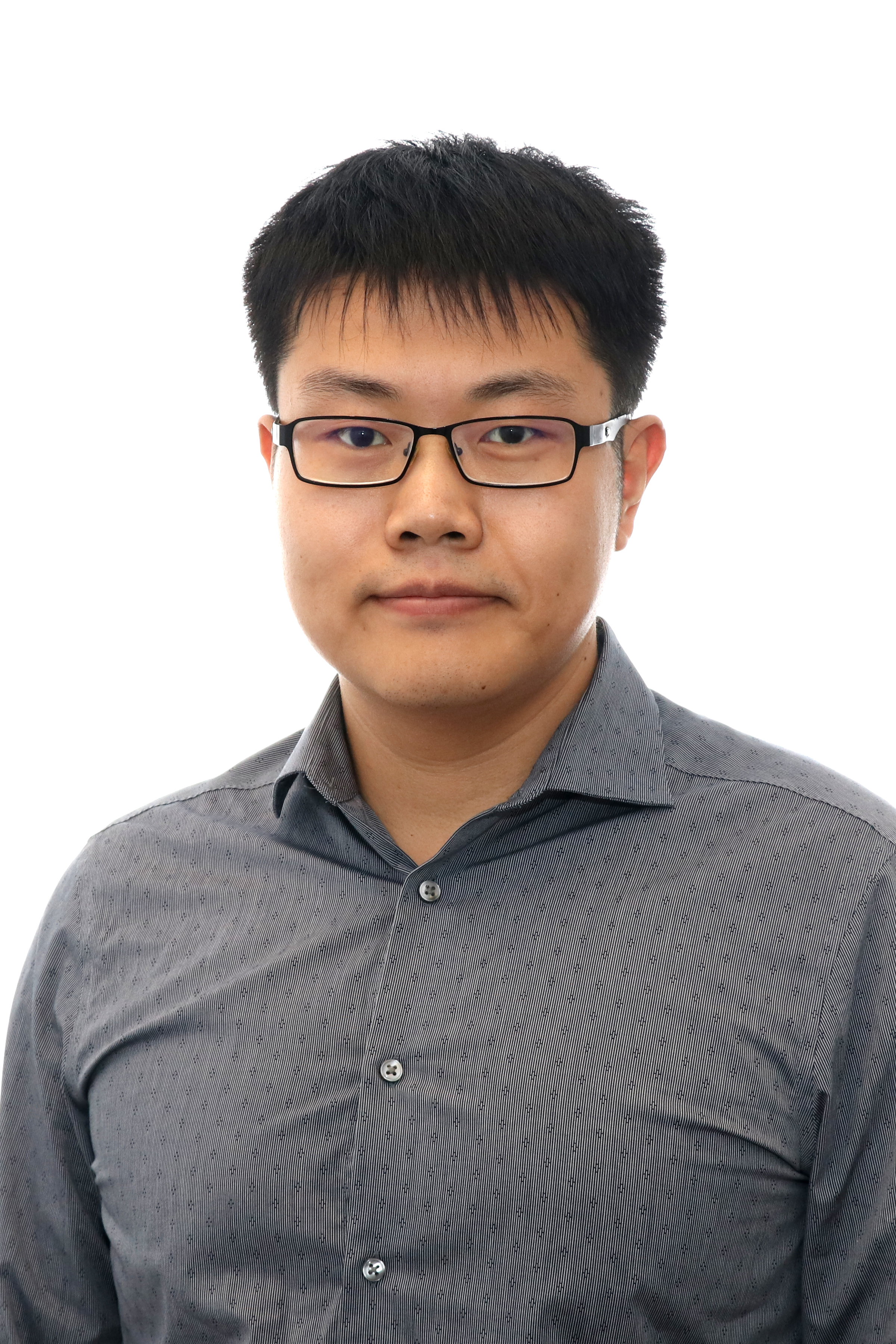}}]
{Hanwang Zhang} is currently an assistant professor at Nanyang Technological University, Singapore. He was a research scientist at the Department of Computer Science, Columbia University, USA. He has received the B.Eng (Hons.) degree in computer science from Zhejiang University, Hangzhou, China, in 2009, and the Ph.D. degree in computer science from the National University of Singapore in 2014. His research interest includes computer vision, multimedia, and social media. Dr. Zhang is the recipient of the Best Demo runner-up award in ACM MM 2012, the Best Student Paper award in ACM MM 2013, and the Best Paper Honorable Mention in ACM SIGIR 2016，and TOMM best paper award 2018. He is also the winner of Best Ph.D. Thesis Award of School of Computing, National University of Singapore, 2014.
\end{IEEEbiography}

\begin{IEEEbiography}[{\includegraphics[width=1in,height=1.25in,clip,keepaspectratio]{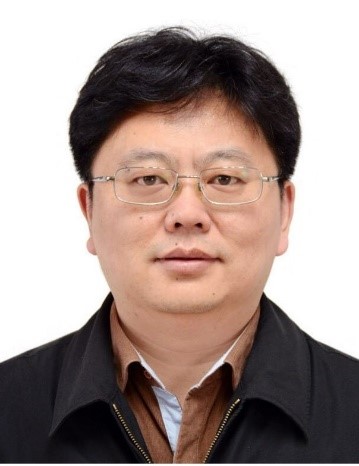}}]
{Yongdong Zhang} (M’08-SM’13) received the Ph.D. degree in electronic engineering from Tianjin University, Tianjin, China, in 2002.He is currently a Professor with the School of Information Science and Technology, University of Science and Technology of China. His current research interests are in the fields of multimedia content analysis and understanding, multimedia content security, video encoding and streaming media technology.
He has authored over 100 refereed journal and conference papers. He was a recipient of the Best Paper Awards in PCM 2013, ICIMCS 2013, and ICME 2010, the Best Paper Candidate in ICME 2011.He serves as an Associate Editor of IEEE Trans. on Multimedia and an Editorial Board Member of Multimedia Systems Journal.
\end{IEEEbiography}

\begin{IEEEbiography}[{\includegraphics[width=1in,height=1.25in,clip,keepaspectratio]{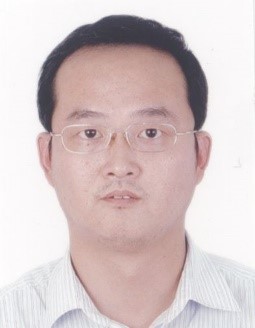}}]
{Feng Wu}(M'99-SM'06-F'13) received the B.S. degree in Electrical Engineering from Xidian University in 1992. He received the M.S. and Ph.D. degrees in Computer Science from Harbin Institute of Technology in 1996 and 1999, respectively. Now he is a professor in University of Science and Technology of China. Before that, he was a principle researcher and research manager with Microsoft Research Asia.
His research interests include computational photography, image and video compression, media communication, and media analysis and synthesis. He has authored or co-authored over 200 high quality papers (including several dozens of IEEE Transaction papers and top conference papers in MOBICOM, SIGIR, CVPR and ACM MM). He has 77 granted US patents. Fifteen of his techniques have been adopted into international video coding standards. As a co-author, he received the best paper award from IEEE T-CSVT 2009, PCM 2008 and SPIE VCIP 2007. Wu has been a Fellow of IEEE. He serves as an associate editor for IEEE Transactions on Circuits and System for Video Technology, IEEE Transactions on Multimedia and several other International journals. He received the IEEE Circuits and Systems Society 2012 Best Associate Editor Award. He also served as the TPC chair for MMSP 2011, VCIP 2010 and PCM 2009, and the Special Sessions chair for ICME 2010 and ISCAS 2013.
\end{IEEEbiography}




\end{document}